\documentclass{article}
\usepackage{enumitem}
\usepackage{graphicx}
\usepackage{xurl}

\PassOptionsToPackage{numbers, compress}{natbib}

 \usepackage[eandd, final]{neurips_2026}

\usepackage[utf8]{inputenc} %
\usepackage[T1]{fontenc}    %
\usepackage{hyperref}       %
\usepackage{url}            %
\usepackage{booktabs}       %
\usepackage{amsmath}        %
\usepackage{amsfonts}       %
\usepackage{amsthm}         %
\usepackage{nicefrac}       %
\usepackage{microtype}      %
\usepackage{xcolor}         %
\usepackage[ruled,noline,linesnumbered]{algorithm2e}

\usepackage{multirow}
\usepackage{placeins}        %

\title{Where Root Cause Analysis Fails: A Retrieval-Reranking Decomposition}

\author{%
  Hada Melino Muhammad\textsuperscript{1} \quad
  Luan Pham\textsuperscript{1} \quad
  Laure Barri\`ere\textsuperscript{2} \\
  \textbf{Sachin Shetty}\textsuperscript{2} \quad
  \textbf{Leonardo Pulga}\textsuperscript{2} \quad
  \textbf{Flora D.~Salim}\textsuperscript{1} \\
  \textsuperscript{1}University of New South Wales \quad
  \textsuperscript{2}Baker Hughes \\
  \texttt{\{hada\_melino.muhammad, luan.pham, flora.salim\}@unsw.edu.au} \\
  \texttt{\{laure.barriere, sachin.shetty, leonardo.pulga\}@bakerhughes.com}
}

\hypersetup{pdftitle={Where Root Cause Analysis Fails: A Retrieval-Reranking Decomposition},
  pdfauthor={Hada Melino Muhammad, Luan Pham, Laure Barri\`ere, Sachin Shetty, Leonardo Pulga, Flora D. Salim}}

\begin{document}

\maketitle

\begin{abstract}
Identifying the root cause of an anomaly among hundreds of sensors is critical for preventing safety incidents and costly downtime in complex monitored systems. Existing studies evaluate root cause analysis (RCA) methods using top@k accuracy. We show that this metric has a fundamental blind spot: it conflates two failure modes, \textit{retrieval failure}, where the true cause is never considered, and \textit{reranking failure}, where it is considered but ranked too low.
In this work, we introduce a retrieval--reranking decomposition and audit four well-known benchmarks to expose this blind spot. Our experiments show that, on industrial CPS benchmarks, the statistical baseline scoring every sensor mis-ranks the true cause 79--100\% of the time, and graph-based methods never clearly beat the best statistical baseline, whether their causal graphs are learned on short fault windows, on retrieved candidate pools guaranteed to contain the cause, or on multi-day normal-operation data. Meanwhile, on microservice benchmarks, where each injected fault drives the faulty service's own metric far above its baseline, retrieval is nearly solved (98--100\%). Guided by the decomposition, we build a two-stage pipeline combining a multi-signal retriever with an LLM reranker that, as one fixed configuration, matches or exceeds the best baseline's top@1 accuracy on all six benchmark suites (by up to $+$12 points), with
no causal graph or labeled data required. When all methods rank the same retrieved candidates with the true cause guaranteed present, adding a short system-description document lets the reranker lead the best baseline by $+$7 to $+$18 points on every benchmark. Code is available at \url{https://github.com/cruiseresearchgroup/DecompRCA}.
\end{abstract}

\section{Introduction} \label{sec:intro}

Complex monitored systems, from water treatment plants~\cite{ahmed2017wadi, goh2016dataset} and 
building management systems~\mbox{\cite{prabowo2024building, OEDI_Dataset_5763}} to cloud service systems~\cite{pham2025rcaeval}, are instrumented with hundreds to thousands of sensors. When
an anomaly or a failure is detected, operators must rapidly identify its root cause to prevent further damage, which may lead to huge financial losses~\cite{yahoo_amazon_downtime_2018} or even fatalities~\cite{Gregory2025Optus}. 
This task is known as root cause analysis (RCA), and automated approaches to it have progressed through statistical analysis~\cite{pham2024baro, shan2019diagnosis}, causal inference-based approaches~\cite{li2022causal, ikram2022root, han2025root}, and LLM-based reasoners~\cite{chen2024automatic}, with each generation reporting improvements on standard benchmarks.
Progress is measured by how well methods surface the true root cause within a short
window of candidates, whether it appears at all and how near the top it lands.
Yet existing benchmarks pre-package the diagnostic context in a way that %
implicitly assumes
the root cause is already exposed: fault scenarios are provided with known injection
timestamps, analysis windows are fixed, and preprocessing decisions are made for the
method rather than by it.
The faults injected in existing microservice benchmarks (e.g., CPU overload) satisfy
this assumption by construction: the faulty service's own metric rises far above its baseline.
This does not generalize: in cyber-physical systems (CPS), faults act through actuators,
setpoints, and control loops, so a root cause can enter its faulty state while its own signal
does not rise above its baseline, and the system is detected through its effects~\cite{li2022causal}.
Even a window split at the labeled fault onset then exposes the effects rather than the cause, so any method evaluated only on
microservice benchmarks inherits a false sense of retrieval coverage that does not transfer
to such systems.\looseness=-1

Every RCA pipeline has, at least implicitly, two stages: a \emph{retriever} that
produces a candidate set $\mathcal{C}$ and a \emph{reranker} that orders $\mathcal{C}$.
The fraction of true causes inside $\mathcal{C}$, which we call Retrieval@$K$ for a set of size $K$, is a hard ceiling on any downstream ranker, since no reranker can recover a
candidate it was never shown, yet existing evaluation collapses both stages into a
single score, making it impossible to determine which stage is responsible for failure.
On CPS, a cause whose own signal does not rise above its baseline is ranked
below its effects by magnitude, and detection order fails because it may be
flagged after them or not at all.
Structure-based methods~\cite{li2022causal, ikram2022root, han2025root} face a further
obstacle: ranking on their learned causal graphs is unreliable, whether the graphs are learned on
short fault windows, on retrieved pools guaranteed to contain the cause, or on multi-day normal
corpora (Section~\ref{sec:results} and Appendix~\ref{app:graphs}), and expert-specified graphs are rarely available in
practice~\cite{ikram2022root}.
The result is a literature where a method that never retrieves the true cause and one
that retrieves but mis-ranks it appear identical under top@1, yet require
categorically different fixes.

We introduce a retrieval--reranking decomposition and apply it as a systematic audit
across four benchmarks (WADI~\cite{ahmed2017wadi}, SWaT~\cite{goh2016dataset},
HVAC~\cite{OEDI_Dataset_5763}, RCAEval~\cite{pham2025rcaeval}) and three method
families.
Critically, WADI and SWaT are among the most widely used industrial benchmarks in the
RCA literature~\cite{zheng2024lemma, han2025root}, yet methods evaluated on them
consistently report low accuracy with no further investigation into why: our
decomposition reveals that this is not an undifferentiated dataset-complexity problem
but two compounding failures: the true cause is often absent from the candidate set
before any ranking is attempted, and even when present it is hard to rank.
To validate that the two failure modes admit independent remedies, we address each
subproblem in isolation: first targeting retrieval failure with a multi-signal retriever
combining deviation magnitude, earliest-onset, and discrete state-change detection;
then targeting reranking failure with a single-call LLM reranker over structured
per-candidate evidence and an optional domain-knowledge document $\mathcal{D}$,
requiring neither a learned causal graph nor labeled fault traces~\cite{ikram2022root}.
The pipeline is meant to validate the decomposition, not to serve as a proposed end
system, and we attribute each of its gains to the stage that produces it.
Our contributions are:\looseness=-1

\begin{itemize}[leftmargin=*,nolistsep]
    \item \textbf{A retrieval--reranking decomposition for RCA evaluation.} We 
    formalize ranking error as two independently measurable failure modes, scored by
    two metrics that apply to any ranking method, Retrieval@$K$ (is the true cause among
    the $K$ candidates?) and Rerank@$k$ (if so, is it ranked among the first $k$?), and
    audit four benchmarks across three method families, surfacing findings invisible
    under top@$k$ alone.\looseness=-1

    \item \textbf{An empirical characterization of the retrieval gap.} Deviation 
    magnitude caps Retrieval@15 at 35--64\% on industrial CPS benchmarks while 
    reaching 98--100\% on microservice benchmarks, with no single additional signal 
    dominating across datasets.

    \item \textbf{Independent remedies for both failure modes.} A multi-signal
    retriever lifts that ceiling to 63--67\%, and an LLM reranker with a
    single configuration for all datasets matches or beats the best baseline's
    top@1 on every benchmark (by up to $+$12 points), showing that each failure
    mode can be fixed on its own. When the true cause is
    guaranteed to be among the retrieved candidates, the reranker is never worse
    than any baseline ranking the same candidates, and a short document describing
    the system lets it beat the best baseline by $+$7 to $+$18 points on every
    benchmark.

    \item \textbf{A design implication.} Anomaly detection and RCA must be 
    co-designed: the anomaly detector determines the retrieval substrate, and 
    misalignment caps accuracy regardless of reranker quality.
\end{itemize}

\section{Related Work} \label{sec:related}

\paragraph{Single-signal RCA scorers.}
Most unsupervised RCA methods attach a deviation score to every
time series and report the resulting order. BARO~\cite{pham2024baro}
proposes a robust median-IQR scorer, $\epsilon$-Diagnosis~\cite{shan2019diagnosis}
ranks by $\epsilon$-statistic, FaaSRCA~\cite{huang2024faasrca} ranks
by graph-autoencoder reconstruction error, KPIRoot+~\cite{gu2026kpiroot}
blends waveform similarity with a recency proxy,
MicroHECL~\cite{liu2021microhecl} prunes a service call graph and
scores the residual nodes by Pearson correlation,
TORAI~\cite{pham2026torai} clusters services by symptom severity
before applying robust hypothesis testing, and
PRISM~\cite{pham2026graph} ranks on an internal-versus-external
property asymmetry that explicitly refutes the maximum-deviation
rule. None forms an explicit candidate set, so top@$k$ merges the two
stages; taking each method's first $K$ items as its candidate set,
Retrieval@$K$ shows whether the cause is buried and Rerank@$k$ whether
it is near the top but misordered (Section~\ref{sec:metrics}). On microservice benchmarks, where
each injected fault drives the faulty service's own metric far above its baseline, this
conflation is benign: the cause is retrieved and
reranking is the binding problem. On industrial CPS benchmarks,
the conflation is costly: the true cause's own signal often does not
rise above its baseline, so a method that scores every sensor can rank it
below any practical candidate budget. Two recent
measurements surface what this conflation costs. Pham
et al.~\cite{pham2024root} report that most causal inference-based
RCA methods barely beat a random baseline at scale, and
TraceDiag~\cite{ding2023tracediag} prunes a 500-node Microsoft
Exchange trace down to about eleven candidates while still retaining
92.9\% of the true causes, in effect a Retrieval@11 of 0.929 reported
without the matching Rerank@$k$. Both findings point the same way: what
enters the candidate set matters as much as how it is reordered, yet no
prior work separates the two.\looseness=-1

\paragraph{Causal graphs and structure-based RCA.}
A second family treats the root cause as the time series occupying
a privileged position in a causal graph. The graph is either
expert-supplied, as in CIRCA~\cite{li2022causal}, which projects
time series onto the four golden signals over a system architecture,
or learned, as in CausalRCA~\cite{xin2023causalrca} (DAG-GNN with
PageRank), RUN~\cite{lin2024root} (neural Granger discovery with
PageRank), CHASE~\cite{zhao2024chase} (causal hypergraph
convolution), and REASON~\cite{wang2023hierarchical} (hierarchical
GNN with random walk with restart, evaluated on WADI and SWaT).
RCD~\cite{ikram2022root} runs a hierarchical PC search,
MicroCERCL~\cite{zhu2024root} adapts the recipe to cloud-edge
deployments, and Cloud Atlas~\cite{xie2024cloud} elicits the graph
itself from an LLM. Across this family, ranking runs end-to-end
over the full metric set; the candidate set is the entire system,
so retrieval failure never separates from reranking failure under
top@$k$. Yet the learned graph is itself a retriever: with its nodes
as the candidate set, $1-$Retrieval@$K$ counts causes lost in graph
learning and Rerank@$k$ scores how graph scoring ranks the rest. Two obstacles make this entanglement
costly. Li
et al.~\cite{li2025root} prove for linear SEMs that propagation amplification can
place downstream metrics above the cause under any
squared-z-score ranker. Learning a reliable graph from a short
non-stationary fault window is also hard, so the Retrieval@$K$ of
structure-based methods tracks graph quality rather than evidence
quality. Nearly all methods in this family are evaluated on
microservice benchmarks where retrieval is trivially saturated; REASON~\cite{wang2023hierarchical} is
the exception, evaluated on SWaT and WADI with a graph learned from
long-run normal data, a setting we also test (Appendix~\ref{app:graphs}).
Our audit runs
these methods under the operationally realistic per-scenario
setting and, for the first time, separates their retrieval failure
from their reranking failure.\looseness=-1

\paragraph{LLM-based RCA.}
A third family applies LLMs as the diagnostic reasoner. One branch
wires multi-agent loops over telemetry: D-Bot~\cite{zhou2024d} runs
a tree-of-thought search across diagnostic tools for database
anomalies, while RCLAgent~\cite{zhang2025adaptive} and
AMER-RCL~\cite{zhang2026agentic} drive trace, log, and metric
agents inside a recursion-of-thought controller. In these systems,
the agent's exploration policy defines a de facto candidate set,
but its retrieval coverage is never measured: a hypothesis the
agent never generates is a retrieval failure indistinguishable from
a mis-ranked hypothesis under top@$k$; logging what the agent
examines would give its Retrieval@$K$. A second branch uses the LLM
as a post-hoc reasoner over a separately-detected candidate set:
KAT~\cite{liu2025kunlun} pairs a kernel-trace anomaly detector with
a fine-tuned 14B Analyzer, and RC-LLM~\cite{zhou2026root} fuses
change-point and call-tree evidence into a single DeepSeek-V3
prompt. In both cases, the detector constitutes an explicit
retrieval stage whose Retrieval@$K$ is never reported separately from
the LLM's Rerank@$k$. Closest to our pipeline is
SpecRCA~\cite{zhang2026hypothesize}, which drafts hypotheses from
multimodal scoring and verifies each in parallel with a distilled
3B model, an architecture that maps directly onto Retrieval@$K$
(the drafter) and Rerank@$k$ (the verifier). Yet SpecRCA, like the others, reports a single
end-to-end accuracy number: a hypothesis missed by the drafter is
silently absorbed into the same accuracy as a hypothesis mis-ranked
by the verifier. None of these systems formalizes
retrieval-versus-reranking failure. Nearly all are evaluated
exclusively on microservice incidents (AIOps, HipsterShop,
Train-Ticket); whether multi-evidence LLM reranking transfers to
industrial cyber-physical fault traces, where retrieval and reranking
fail together, is the question this paper addresses.

\paragraph{Listwise LLM reranking.}
Once a candidate set is available, the reranking problem becomes:
given a small set of sensors with heterogeneous evidence, which is
most likely the root cause? This is structurally analogous to the
listwise reranking problem in information retrieval, where a short
candidate list must be ordered by relevance. RankGPT-style listwise
prompting~\cite{ma2023zero} showed that placing a small candidate
set in one LLM call outperforms pointwise scoring; subsequent work
strengthens the design along orthogonal axes.
Rank-without-GPT~\cite{zhang2025rank} ports the recipe to
open-source backbones, FIRST~\cite{reddy2024first} cuts inference
cost by reading first-token logits, Rank-R1~\cite{zhuang2025rank}
adds RL-trained reasoning, Rank-K~\cite{yang2025rank} distills
chain-of-thought traces into a 32B listwise model, and
AcuRank~\cite{yoon2025acurank} budgets LLM calls under uncertainty.
LLMs are a natural fit for fault evidence reranking for three
reasons. First, the appropriate weighting of heterogeneous evidence
types is context-dependent and varies across fault types and system
architectures; a fixed scoring function cannot adapt to this
variation, but an LLM can reason over the specific pattern present
in each fault instance. Second, LLMs can consume optional
natural-language domain knowledge without requiring it to be
formalized as a causal graph or structured model, lowering the
barrier for operational deployment. Third, the listwise paradigm
is efficient when the candidate set is small: one call per fault,
with no causal graph to learn per scenario.
We adopt this
paradigm to validate that reranking failure is independently
addressable once the retrieval problem is controlled.

\paragraph{Industrial CPS anomaly detection and benchmarks.}
Our evaluation builds on a long line of cyber-physical anomaly
detectors. Early SWaT baselines compare LSTM autoencoders with
one-class SVM~\cite{inoue2017anomaly} and architecture-searched
MLPs trained by genetic algorithm~\cite{shalyga2018anomaly}, both
of which produce a deviation-based ranked list of suspect tags.
GiBy~\cite{venugopalan2026giantstepbabystepclassifierscalable}
couples per-sensor bounds with discrete actuator-state lookups,
and STOD~\cite{wang2020defending} adds spatio-temporal graph
reasoning. These detectors sit upstream of root cause ranking and
motivate the discrete state-change signal we add at retrieval time;
their ranked suspect lists can be scored directly with Retrieval@$K$,
a criterion for choosing a detector to pair with RCA. On the
benchmark side, RCAEval~\cite{pham2025rcaeval} and
PetShop~\cite{hardt2023petshop} provide labeled microservice failure
traces and evaluation protocols, with PetShop reporting that methods
strong on one benchmark frequently fail on another, a transfer
failure our decomposition explains structurally: methods optimized
for microservice benchmarks inherit retrieval assumptions that do not
hold on industrial CPS benchmarks. LEMMA-RCA~\cite{zheng2024lemma}
evaluates six methods on both microservice and operational-technology (OT) datasets
(including SWaT and WADI), observing substantially worse
performance on OT benchmarks, but attributes this to dataset
complexity rather than the retrieval--reranking structure. We adopt
RCAEval as our saturated microservice benchmark and pair it with
WADI~\cite{ahmed2017wadi}, SWaT~\cite{goh2016dataset}, and an HVAC
building-management testbed~\cite{OEDI_Dataset_5763} to measure the
retrieval ceiling on industrial CPS that prior work
has not diagnosed.

\section{Problem Formulation}
\label{sec:motivation}

\begin{figure}
    \centering
    \includegraphics[width=\linewidth]{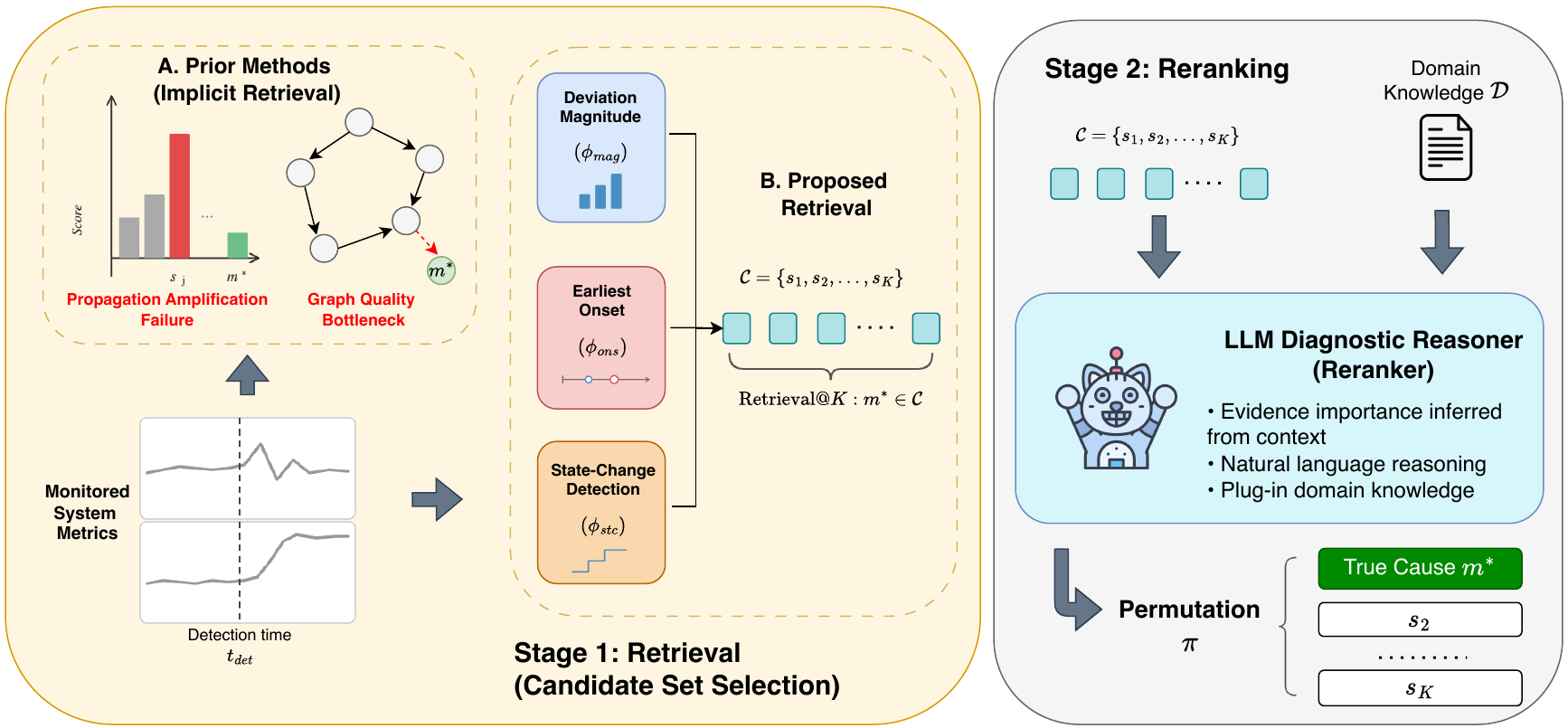}
    \caption{\textbf{Our retrieval--reranking decomposition for RCA.} Left: prior methods fail via mis-ranking (statistical) or graph quality bottlenecks (graph-based). Stage~1 retrieves a candidate set $\mathcal{C}$ using three complementary signals. Stage~2 reranks $\mathcal{C}$ with an LLM reasoner conditioned on per-candidate evidence and optional domain knowledge $\mathcal{D}$, outputting permutation $\pi$.}
    \label{fig:overview}
\end{figure}

Consider a monitored system with $p$ observable metrics. Let
$\mathbf{X}_B \in \mathbb{R}^{T_B \times p}$ denote a \emph{baseline
window} collected under normal operation and
$\mathbf{X}_F \in \mathbb{R}^{T_F \times p}$ a \emph{fault window}
collected after $t_{\mathrm{det}}$, the detection
timestamp delimiting them. An optional domain-knowledge document
$\mathcal{D}$ describing system architecture, component roles, and
causal dependencies may be provided. The goal is to produce a
permutation $\pi \in \mathrm{Sym}(\{1,\ldots,p\})$ ranking metrics
from most to least likely root cause. This setup
pre-packages the diagnostic context: the fault window, baseline
window, and detection timestamp are all assumed to expose the root
cause, whose own signal rises above its baseline inside the fault window. Every
method evaluated under this setup inherits the assumption without stating it:
when it holds, as on microservice benchmarks, top@$k$ measures ranking alone,
and when it fails, a cause the window never exposed is scored as one more
ranking miss. Current benchmarks thus evaluate how well a method ranks within
a window prepared for it, not whether that window exposes the cause.

\section{Decomposition of Root Cause Analysis}
\label{sec:decomposition}

\subsection{Why Root Cause Identification Reduces to Retrieval and Reranking}
\label{sec:rca_formulation}

\paragraph{The ideal case.}
Write the $p$ sensors as $\mathcal{S} =
\{s_1, \ldots, s_p\}$, and let $m^*$ denote the true root cause. In the ideal case, an oracle anomaly
detector observes the true fault onset time $t_{\mathrm{fault}}$
and produces a ranking of sensors by temporal precedence. If $m^*
\in \mathcal{S}$ and the detector is noiseless, then $m^*$ is the
first to turn abnormal, at $t_{\mathrm{fault}}$, and RCA reduces
to a precedence problem: rank sensors by onset and return the
earliest. Formally, let $\tau(s_i)$ denote the time the component
behind $s_i$ enters an abnormal state, whether or not its signal shows
it, so $t_{\mathrm{fault}} = \tau(m^*) \leq \tau(s_j)$ for every $j$. Then:
\begin{equation}
    m^* = \arg\min_{s_i \in \mathcal{S}} \tau(s_i)
\end{equation}
and the problem is solved by precedence alone. No reranking is
required.

\paragraph{The practical case: noisy detectors confound precedence.}
In practice, no oracle provides $t_{\mathrm{fault}}$. Anomaly
detectors observe $t_{\mathrm{det}} = \min_i t_{\mathrm{det}}(s_i)$, when
the system is first \emph{detected} as faulty, a noisy, delayed proxy for
$t_{\mathrm{fault}}$; $t_{\mathrm{det}}(s_i) = \tau(s_i) + \delta_i$ is when
the signal of $s_i$ is flagged, with detection lag $\delta_i \geq 0$.
Because $\delta_i$ varies across sensors depending on detection
sensitivity, signal-to-noise ratio, and propagation gain, the
ordering of detected anomalies need not preserve the ordering of
true onsets:
\begin{equation}
    \label{eq:precedence}
    t_{\mathrm{det}}(s_i) < t_{\mathrm{det}}(s_j)
    \not\Rightarrow \tau(s_i) < \tau(s_j)
\end{equation}
A downstream sensor $s_j$ that responds abruptly to the propagating
fault may be \emph{detected} before the root cause $m^*$ despite
having a later true onset. Nor is $\delta_i$ bounded: if the cause's own
signal does not rise above its baseline (a held or switched actuator, a
loop-compensated sensor bias), a detector that looks for rises lags on it: $\delta_{m^*}$ can exceed the window, so
$t_{\mathrm{det}}(m^*) > t_{\mathrm{det}}$ or $m^*$ is never flagged. The
cause is then \emph{hidden} from magnitude and unexposed by the window: the system is detected through its effects,
which both detection order and magnitude favor. Ranking by precedence therefore fails
not because precedence is the wrong signal, but because the
available signals are $t_{\mathrm{det}}(s_i)$ rather than $\tau(s_i)$.\looseness=-1

\paragraph{How prior methods embed implicit retrieval assumptions.}
Given noisy detectors, prior methods reduce the ranking problem
by committing to a signal $\phi: \mathcal{S} \rightarrow \mathbb{R}$
that scores each sensor and implicitly defines a candidate set
$\mathcal{C} = \text{top-}K(\phi)$.

\emph{Statistical methods} that score every sensor, with $\phi$ a
deviation magnitude or precedence statistic,
have $\mathcal{C} = \mathcal{S}$. Retrieval failure is zero by
construction, but $\phi$ mis-ranks a cause hidden from it.

\emph{Graph-based methods} define a function $F: \mathbf{X}_F,
\mathbf{X}_B \rightarrow \mathcal{G}(\mathcal{N}, \mathcal{E})$
that recovers a causal graph, then set $\mathcal{C}$ to the nodes
occupying a privileged position in $\mathcal{G}$. If the true
causal path is $m^* \rightarrow a \rightarrow b \rightarrow \cdots
\rightarrow c$, a missing edge anywhere in the path can exclude
$m^*$ from $\mathcal{C}$ entirely; e.g., the first edge:
\begin{equation}
    (m^*, a) \notin \mathcal{E} \Rightarrow m^* \notin \mathcal{C}
\end{equation}
Graph methods can therefore fail at retrieval before reranking is
attempted, in addition to failing at ranking the nodes the graph keeps;
where the two can be separated, Section~\ref{sec:results} shows that the second failure
dominates.\looseness=-1

\paragraph{The role of domain knowledge.}
Since $\tau$ is unobservable in practice, the
precedence-based ideal solution is inaccessible. However, if $m^* \in
\mathcal{C}$ can be guaranteed by the retriever, the reranking
problem becomes: given heterogeneous evidence for each candidate in
$\mathcal{C}$, which is most likely the origin of the fault? This
is where domain knowledge $\mathcal{D}$, describing system
architecture, component roles, and causal dependencies, can be
injected. Formally, the reranker computes:
\begin{equation}
    \hat{m}^* = \arg\max_{s_i \in \mathcal{C}}\,
    \mathcal{F}(s_i \mid \mathbf{X}_F, \mathbf{X}_B, \mathcal{D})
\end{equation}
where $\mathcal{F}$ is a scoring function that integrates
statistical evidence with prior knowledge $\mathcal{D}$. Unlike
graph-based methods that require $\mathcal{D}$ to be formalized
as a causal graph, a flexible reranker can consume $\mathcal{D}$
in natural language form, conditioning its ranking on system
context without committing to a fixed structural model. The
decomposition therefore implies a design principle: retrieval
should maximize $\mathbf{1}[m^* \in \mathcal{C}]$ using observable
signals, and reranking should maximize
$\mathbf{1}[\hat{m}^* = m^* \mid m^* \in \mathcal{C}]$ using
all available evidence including $\mathcal{D}$.\looseness=-1

\section{A Two-Stage Remedy: Retrieval Then Reranking}
\label{sec:remedy}

The decomposition makes the two failure modes independently addressable. We
instantiate a two-stage pipeline (Figure~\ref{fig:overview}) as a validation
instrument: closing each gap requires categorically different interventions,
and progress on one does not substitute for the other.

\subsection{Stage 1: Multi-Signal Retrieval}

Retrieval produces a candidate set $\mathcal{C}$ of at most $K$ sensors that maximizes
$\mathbf{1}[m^* \in \mathcal{C}]$. Magnitude alone can miss a cause
hidden from it (Section~\ref{sec:rca_formulation}), so we combine three
complementary signals.

\textbf{Deviation magnitude} $\phi_{\text{mag}}$: largest upward normalized shift of each
sensor from its baseline window. Effective when the cause's own signal
rises far above its baseline (e.g., resource-stress faults).

\textbf{Earliest onset} $\phi_{\text{ons}}$: each sensor's first-deviation
timestamp within the fault window, earliest first. Recovers the precedence
signal of Section~\ref{sec:rca_formulation} when the cause deviates early
but weakly or downward.

\textbf{Discrete state-change} $\phi_{\text{stc}}$: whether a discrete-valued
sensor leaves its baseline mode, ranked by when it first does. Targets binary actuator or mode-switch
faults whose signature is a discrete event rather than a continuous shift
(SWaT-style abrupt transitions).

The budget $K$ is split evenly across the signals unless noted, and the
deduplicated union forms $\mathcal{C}$ ($|\mathcal{C}| \leq K$).
These three signals are illustrative, not a fixed prescription: each targets a
regime we observed in our benchmarks (Section~\ref{sec:results}), and a system that surfaces faults
through a different signature (e.g., spectral shifts, correlation breaks)
would warrant adding the corresponding scorer.

\subsection{Stage 2: LLM Reranking}

Given $\mathcal{C}$, the reranker answers a different question: which candidate
is most likely the true cause? For each $s_i \in \mathcal{C}$ we build an
evidence record (deviation magnitude, time of first onset, baseline and
fault-window means with absolute and relative shift), pass all records in a
single prompt to an LLM, and read off the returned permutation $\pi$. An optional domain-knowledge document $\mathcal{D}$
describing system architecture and propagation pathways is injected into the
system prompt; unlike graph-based methods, the reranker accepts $\mathcal{D}$
in informal natural language without requiring formalization as a causal
graph. The pipeline is unsupervised, uses no labeled fault traces, and runs
a single configuration across all datasets (Appendix~\ref{app:reproducibility}).\looseness=-1

\section{Experimental Setup}
\label{sec:experiments}

\subsection{Datasets}
We evaluate on four operational benchmarks spanning two distinct
domains. \textbf{RCAEval}~\cite{pham2025rcaeval} is a microservice
benchmark with three suites, Online Boutique (RE1-OB), Sock Shop
(RE1-SS), and Train Ticket (RE1-TT), each with $n{=}125$ scenarios and
evaluated at the service level. \textbf{WADI}~\cite{ahmed2017wadi}
($n{=}14$), \textbf{SWaT}~\cite{goh2016dataset} ($n{=}36$), and
\textbf{HVAC}~\cite{OEDI_Dataset_5763} ($n{=}48$) are industrial CPS
benchmarks covering water distribution, water treatment, and building
management respectively, evaluated at the metric level. The four
datasets differ in system scale, fault type, sampling rate, and
ground-truth granularity, providing a diverse testbed for evaluating
retrieval and reranking across microservice and industrial CPS
benchmarks. Full dataset details are provided in
Appendix~\ref{app:datasets}.\looseness=-1

\subsection{Baselines}
We compare against two families of baselines. \emph{Statistical
baselines} score each sensor directly, without a learned global graph:
BARO~\cite{pham2024baro}, RCD~\cite{ikram2022root} (which runs a
localized causal search), and $\epsilon$-Diagnosis~\cite{shan2019diagnosis}.
\emph{Graph-based baselines} learn a causal graph per scenario via PC or
FCI~\cite{spirtes2000causation} and rank sensors on it with
PageRank~\cite{Page1998PageRank}, random walk~\cite{tong2006fast}, or
CIRCA~\cite{li2022causal}. All baselines use their published default
hyperparameters~\cite{pham2025rcaeval}, with dataset-specific patch sizes
where the windows require it, and score the same baseline and fault
windows as our pipeline (Appendix~\ref{app:baselines}).

\paragraph{LLM reranker configurations.}
The reranker is \texttt{gpt-oss-120b}, run $n{=}3$ times at $T{=}1.0$
(we report mean$\pm$std) with the same prompt on every dataset
(Appendix~\ref{app:reproducibility}). We report it both without and
with the domain-knowledge (DK) document $\mathcal{D}$ (no-DK / with-DK) on every dataset,
and use $K{=}15$ throughout.\looseness=-1

\subsection{Evaluation Metrics}
\label{sec:metrics}
We report \textbf{top@$k$} (1 if any ground-truth cause appears
among the first $k$ items of the final ranking, averaged over scenarios) for
$k \in \{1,3,5\}$ and \textbf{Avg@5}
($= \frac{1}{5}\sum_{k=1}^{5}$\,top@$k$) for end-to-end ranking
evaluation. To separate the two stages we define two metrics over a
candidate set $\mathcal{C}$ of size $K$. \textbf{Retrieval@$K$} is the
fraction of scenarios whose ground-truth cause appears in
$\mathcal{C}$; \textbf{Rerank@$k$} is, among those scenarios, the
fraction in which the cause is ranked among the first $k$; $K$ is the
candidate-set size and $k \leq K$ the cutoff on the final ranking. By
construction, top@$k$ $=$ Retrieval@$K$ $\times$ Rerank@$k$, so
$1 -$ Retrieval@$K$ is the retrieval failure and Retrieval@$K$ $-$
top@$k$ the reranking failure. Both apply to any method that outputs
a ranking: take $\mathcal{C}$ to be the first $K$ items of its final
ranking, so Retrieval@$K$ $=$ top@$K$ and Rerank@$k$ $=$
top@$k$ $/$ top@$K$. For a two-stage pipeline whose reranker only
reorders the retrieved set, this $\mathcal{C}$ is exactly the
retrieved set. We use $K \in \{5, 10, 15\}$ and report $K$ alongside
both metrics.\looseness=-1

\section{Results}
\label{sec:results}

\subsection{Retrieval Failure}

\begin{table}[t]
  \caption{Cumulative-budget Retrieval@$K$. At each $K$, the
  candidate pool is split across signals while the total pool size is
  held fixed: \textbf{mag} uses $K$ magnitude items; \textbf{+ons}
  splits $K$ between magnitude and onset; \textbf{+stc} splits it across
  magnitude, onset, and state-change (Appendix~\ref{app:reproducibility}). Best per
  (dataset, $K$) in \textbf{bold}.}
  \label{tab:retrieval-recall}
  \centering
  \footnotesize
  \setlength{\tabcolsep}{4pt}
  \begin{tabular}{l ccc ccc ccc}
    \toprule
    & \multicolumn{3}{c}{$K=5$}
    & \multicolumn{3}{c}{$K=10$}
    & \multicolumn{3}{c}{$K=15$} \\
    \cmidrule(lr){2-4}\cmidrule(lr){5-7}\cmidrule(lr){8-10}
    Dataset & mag & +ons & +stc & mag & +ons & +stc & mag & +ons & +stc \\
    \midrule
    \multicolumn{10}{l}{\textit{Industrial CPS benchmarks}} \\
    \midrule
    WADI ($n{=}14$)
      & 0.36 & 0.36 & \textbf{0.50}
      & 0.43 & 0.50 & \textbf{0.57}
      & \textbf{0.64} & 0.57 & \textbf{0.64} \\
    SWaT ($n{=}36$)
      & \textbf{0.33} & 0.25 & 0.31
      & 0.44 & 0.44 & \textbf{0.61}
      & 0.53 & 0.61 & \textbf{0.67} \\
    HVAC ($n{=}48$)
      & 0.21 & 0.06 & \textbf{0.25}
      & 0.27 & \textbf{0.56} & 0.54
      & 0.35 & \textbf{0.65} & 0.63 \\
    \midrule
    \multicolumn{10}{l}{\textit{Microservice benchmarks (RCAEval)}} \\
    \midrule
    RE1-OB ($n{=}125$)
      & \textbf{0.98} & 0.95 & 0.94
      & \textbf{1.00} & 0.98 & 0.96
      & \textbf{1.00} & 0.99 & 0.98 \\
    RE1-SS ($n{=}125$)
      & \textbf{1.00} & 0.99 & 0.96
      & \textbf{1.00} & \textbf{1.00} & 0.99
      & \textbf{1.00} & \textbf{1.00} & \textbf{1.00} \\
    RE1-TT ($n{=}125$)
      & \textbf{0.91} & 0.83 & 0.74
      & \textbf{0.98} & 0.91 & 0.90
      & \textbf{0.98} & \textbf{0.98} & 0.91 \\
    \bottomrule
  \end{tabular}
\end{table}

Table~\ref{tab:retrieval-recall} reveals a structural divide between the two
benchmark families. On microservice benchmarks, deviation magnitude alone achieves
Retrieval@15 of 0.98--1.00, confirming that retrieval is trivially solved when
each fault drives the faulty service's own metric far above its baseline. On industrial CPS benchmarks, magnitude
alone caps Retrieval@15 at 0.35--0.64, meaning the true cause is absent from
the candidate set in over a third of faults before any reranking is attempted;
in 43 of these 53 misses the cause's own signal does not rise above its baseline: it switches
state, drops, or does not change.\looseness=-1

Adding complementary signals narrows this gap in a dataset-dependent way. On HVAC,
the onset lift ($0.354 \to 0.646$, $+29$ percentage points, pp) depends on the
order of tied onsets and vanishes if they are ranked by
deviation size (0.31). On SWaT, onset's $K{=}15$ gain holds for any tie order, and discrete state-change detection gives the largest lift
($0.44 \to 0.61$ at $K{=}10$, where onset adds nothing), as binary actuator faults produce abrupt transitions that magnitude
under-weights. On WADI, the added signals help at small budgets ($K{=}5, 10$)
but not at $K{=}15$, where magnitude already reaches 0.64. On microservice benchmarks, adding signals never raises and often \emph{reduces} Retrieval@$K$,
since displacing magnitude items that already cover the cause adds noise
without benefit.

\paragraph{Three saturation regimes.}
Figure~\ref{fig:saturation} (Appendix~\ref{app:full-grids}) allocates each signal its own full budget $K$,
rather than splitting one shared budget, and reveals three qualitatively
distinct retrieval regimes. \textbf{Amplitude-suppressed} (HVAC): magnitude
reaches only 0.60 even at $K{=}50$, and onset's 0.67 by $K{=}10$ reflects
the column order of tied onsets. \textbf{Rank-displaced}
(WADI, SWaT): the cause is outranked at $K{=}10$--$20$ and reached by magnitude
only at large $K$. On SWaT, onset recovers it at a budget up to 20 smaller via
temporal precedence, while on WADI its lead in the figure does not carry over to
the shared budget of Table~\ref{tab:retrieval-recall} (0.57 vs.\ 0.64 at $K{=}15$).
\textbf{Magnitude-dominant} (RCAEval): each fault drives the faulty service's own
metric far above its baseline; onset and state-change add little.
Which signal is informative is thus a property of the system, and the detector
that defines the fault window must be co-designed with retrieval.

\subsection{Reranking Failure}

\begin{table}[t]
  \caption{Headroom decomposition for the multi-evidence LLM reranker
  without the domain document (no-DK; with-DK results in
  Table~\ref{tab:main-results}).
  Retrieval failure $= 1 -$ Retrieval@15; reranking failure $=$
  Retrieval@15 $-$ top@1; Rerank@1 $=$ top@1 $/$ Retrieval@15, best
  per group in \textbf{bold}.}
  \label{tab:headroom}
  \centering
  \footnotesize
  \setlength{\tabcolsep}{5pt}
  \begin{tabular}{lcccccc}
    \toprule
    & BARO  & Retrieval@15 & LLM (no DK) & Retrieval & Reranking & \\
    Dataset & top@1 & (+stc, $K{=}15$) & top@1 & failure & failure & Rerank@1 \\
    \midrule
    \multicolumn{7}{l}{\textit{Industrial CPS benchmarks}} \\
    \midrule
    WADI ($n{=}14$)
      & 0.214 & 0.64 & 0.333$\pm$0.082 & 0.36 & 0.31 & \textbf{0.52} \\
    SWaT ($n{=}36$)
      & 0.194 & 0.67 & 0.213$\pm$0.016 & 0.33 & 0.45 & 0.32 \\
    HVAC ($n{=}48$)
      & 0.000 & 0.63 & 0.153$\pm$0.043 & 0.38 & 0.47 & 0.24 \\
    \midrule
    \multicolumn{7}{l}{\textit{Microservice benchmarks (RCAEval)}} \\
    \midrule
    RE1-OB ($n{=}125$)
      & 0.784 & 0.98 & 0.875$\pm$0.009 & 0.02 & 0.10 & \textbf{0.90} \\
    RE1-SS ($n{=}125$)
      & 0.856 & 1.00 & 0.872$\pm$0.024 & 0.00 & 0.13 & 0.87 \\
    RE1-TT ($n{=}125$)
      & 0.560 & 0.91 & 0.653$\pm$0.024 & 0.09 & 0.26 & 0.72 \\
    \bottomrule
  \end{tabular}
\end{table}

Table~\ref{tab:headroom} separates residual error into retrieval and
reranking components. On microservice benchmarks, retrieval failure
is near zero (0.00--0.09) and Rerank@1 is high
(0.72--0.90), confirming that the dominant bottleneck is reranking
rather than retrieval. On industrial CPS benchmarks, both failure
modes are active simultaneously: retrieval failure accounts for
33--38\% of scenarios, while Rerank@1 is highly
dataset-dependent (0.24--0.52). The gap between microservice and CPS
Rerank@1 (0.72--0.90 vs.\ 0.24--0.52) cannot be attributed
to retrieval alone; even among scenarios where the true cause is
retrieved, the LLM reranker places it first far less reliably on CPS.

BARO scores every sensor, so at $K{=}p$ its Retrieval@$K$ is 1 apart from
the two unobservable causes (Appendix~\ref{app:datasets}), yet its
top@1 on CPS is at most $0.214$: its failure is almost entirely at
reranking, with the true cause not ranked first 79--100\% of the time. The two gaps are thus separate problems with
separate fixes: closing one leaves the other as the binding constraint.

\subsection{Addressing Each Failure Mode}

Table~\ref{tab:main-results} reports end-to-end ranking accuracy
without and with the domain document (full top@1/3/5/Avg@5 grids in
Appendix~\ref{app:full-grids}). Without the document, the multi-evidence LLM reranker matches or exceeds all
statistical and graph-based baselines on top@1 across all datasets, by
up to $+12$ pp on industrial CPS (WADI) and $+9$ pp on microservices
(RE1-TT); no baseline beats it anywhere at top@1. Margins are narrow
on SWaT and HVAC, where a third of causes never reach the reranker,
and on RE1-SS, where BARO already reaches $0.856$.

On industrial CPS, no graph-based method beats the best statistical baseline. CIRCA,
PageRank, and RandomWalk over per-scenario PC/FCI graphs all achieve
top@1 $\leq 0.214$ across the three CPS benchmarks (best graph row
$0.214$ on WADI, tying BARO; $0.111$ on SWaT and $0.104$ on HVAC, below
BARO's $0.194$ and $\epsilon$-Diagnosis's $0.146$). The decomposition explains why: per-scenario
graph construction drops the ground-truth cause in some scenarios
(its near-constant and collinearity filters remove it in 1/14 WADI,
2/36 SWaT, and 12/48 HVAC scenarios; retrieval failure), and graph-based scoring (whether unweighted centrality or CIRCA's
regression-based hypothesis test) correlates poorly with the
manipulated sensor on the surviving pool (reranking failure). Both
failure modes compound, leaving graph-based methods no better than
the best statistical baseline despite their additional structural machinery.
Nor is it a matter of graph size or window length: graphs fitted on
${\leq}15$-node pools that provably contain the cause still do not clearly beat
the best statistical baseline on CPS (best graph row $0.231$ vs.\ BARO $0.308$ on
WADI, $0.083$ vs.\ RCD $0.188$ on HVAC, and $0.200$, tying BARO, on
SWaT) while lifting CIRCA on microservices (RE1-OB
$0.576 \to 0.720$), and global graphs learned from multi-day
normal-operation corpora do not help either (top@1 $\leq 0.214$ on
CPS, no better than per-scenario graphs). Their residual failure is reranking-dominated: the cause is in
the learned graph in 71.4\% (WADI) and 86.1\% (SWaT) of scenarios but
ranked first in at most 21.4\% and 8.3\% (Appendix~\ref{app:graphs}).\looseness=-1

The effect of DK is dataset-dependent and
non-monotone end to end. On HVAC,
DK improves top@1 substantially ($0.153 \to 0.278$, $+12.5$ pp),
suggesting that the document supplies process structure the evidence
records lack (Appendix~\ref{app:robustness}). On SWaT, DK \emph{hurts} top@1 ($-6.5$ pp); on WADI the
top@1 change is within run-to-run noise ($-2.4$ pp on top@1, but
$+9.5$ and $+11.9$ pp on top@3 and top@5). On microservices, DK
gives consistent modest gains (RE1-OB: $+1.3$; RE1-SS: $+7.2$;
RE1-TT: $+3.2$ pp top@1). Section~\ref{sec:controlled} shows that on SWaT,
DK's benefit falls on causes the retriever misses, so end to end only its
cost on retrieved causes is visible.\looseness=-1

\begin{table}[t]
  \caption{End-to-end main results (top@1). LLM rows:
  mean$\pm$std over $n{=}3$ runs at $T{=}1.0$, without and with the
  domain document; other rows single runs. HVAC LLM rows count the
  causes the retriever misses as misses
  (Appendix~\ref{app:hvac-correction}). Full top@1/3/5/Avg@5 grids
  in Appendix~\ref{app:full-grids}.}
  \label{tab:main-results}
  \centering
  \footnotesize
  \setlength{\tabcolsep}{3.5pt}
  \begin{tabular}{l ccc ccc}
    \toprule
    & \multicolumn{3}{c}{Industrial CPS} & \multicolumn{3}{c}{Microservices (RCAEval)} \\
    \cmidrule(lr){2-4}\cmidrule(lr){5-7}
    Method & WADI & SWaT & HVAC & RE1-OB & RE1-SS & RE1-TT \\
    \midrule
    BARO                 & 0.214 & 0.194 & 0.000 & 0.784 & 0.856 & 0.560 \\
    RCD                  & 0.071 & 0.083 & 0.083 & 0.304 & 0.248 & 0.144 \\
    $\epsilon$-Diagnosis & 0.000 & 0.028 & 0.146 & 0.072 & 0.232 & 0.008 \\
    CIRCA (PC)           & 0.214 & 0.083 & 0.062 & 0.536 & 0.632 & 0.320 \\
    CIRCA (FCI)          & 0.214 & 0.111 & 0.104 & 0.576 & 0.624 & 0.328 \\
    PageRank (PC)        & 0.000 & 0.028 & 0.021 & 0.112 & 0.136 & 0.008 \\
    PageRank (FCI)       & 0.071 & 0.000 & 0.000 & 0.096 & 0.184 & 0.048 \\
    RandomWalk (PC)      & 0.071 & 0.000 & 0.000 & 0.080 & 0.112 & 0.064 \\
    RandomWalk (FCI)     & 0.000 & 0.000 & 0.000 & 0.112 & 0.144 & 0.072 \\
    \midrule
    LLM (no DK)
      & \textbf{0.333}$\pm$0.082 & \textbf{0.213}$\pm$0.016 & 0.153$\pm$0.043
      & 0.875$\pm$0.009 & 0.872$\pm$0.024 & 0.653$\pm$0.024 \\
    LLM (with DK)
      & 0.310$\pm$0.041 & 0.148$\pm$0.016 & \textbf{0.278}$\pm$0.012
      & \textbf{0.888}$\pm$0.008 & \textbf{0.944}$\pm$0.008 & \textbf{0.685}$\pm$0.009 \\
    \bottomrule
  \end{tabular}
\end{table}

\subsection{Controlled Comparisons and Robustness}
\label{sec:controlled}

Because the LLM ranks the retriever's candidates while the baselines
rank all sensors, we also compare all methods on the same candidates
(Appendix~\ref{app:controls}).\looseness=-1

\paragraph{Same-candidate controls.}
Reranking the LLM's exact $K{=}15$ candidate lists by each signal
alone and by an equal-weight Borda fusion of all three shows that no fixed rule
generalizes: the best single rule changes across
benchmarks (magnitude on microservices, onset on SWaT, state-change
on HVAC), Borda never wins one outright, and the rule that wins HVAC scores $0.000$ on RE1-TT. The
LLM, as one uniform configuration, beats or ties every rule and
fusion on five of six benchmarks and overtakes the sixth (HVAC) once
domain knowledge is added ($0.326$ vs.\ $0.250$).

\paragraph{Retrieval-controlled pools.}
With Retrieval@$K$ pinned to 1 (a reserved spot for the true cause) and
all nine baselines re-run on the identical pools
(Table~\ref{tab:recall-controlled}), the LLM is never worse than any
baseline, with one tie (HVAC retriever pool, $0.188 =$ RCD; Appendix~\ref{app:controls}
bounds the effect of where the reserved cause is placed). With domain
knowledge, which no baseline can consume, it leads the best
same-pool baseline on every benchmark, by $+6.7$ to $+18.1$~pp
(Table~\ref{tab:recall-controlled}, $\Delta$). The all-candidates column,
run without DK like the baselines, varies only the pool size, and its
effect is dataset-dependent: on WADI and SWaT the
LLM gets \emph{worse} than at $K{=}15$ (WADI $0.410 \to 0.385$, SWaT
$0.229 \to 0.210$; plausibly as structured distractors), while
on HVAC and the microservices it is benign to helpful (HVAC
$0.188 \to 0.299$, RE1-OB $0.883 \to 0.896$, RE1-TT
$0.688 \to 0.795$).

\begin{table}[t]
  \caption{Retrieval-controlled top@1 (true cause present in every
  pool). BB $=$ best of nine baselines on identical pools; $\Delta$ $=$
  best LLM column minus BB, in pp. WADI/SWaT $n{=}13/35$
  (causes outside the evaluated sensor set cannot be pinned). Full grids in
  Appendix~\ref{app:controls}.}
  \label{tab:recall-controlled}
  \centering
  \footnotesize
  \setlength{\tabcolsep}{4.5pt}
  \begin{tabular}{lcccc@{\hspace{10pt}}ccc}
    \toprule
    & \multicolumn{4}{c}{Retriever pool ($K{=}15$)}
    & \multicolumn{3}{c}{All candidates} \\
    \cmidrule(lr){2-5}\cmidrule(lr){6-8}
    Group & BB & LLM (no DK) & LLM (with DK) & $\Delta$ & BB & LLM (no DK) & $\Delta$ \\
    \midrule
    WADI   & 0.308 & 0.410$\pm$0.044 & \textbf{0.436}$\pm$0.089 & $+$12.8 & 0.231 & \textbf{0.385}$\pm$0.077 & $+$15.4 \\
    SWaT   & 0.200 & 0.229$\pm$0.029 & \textbf{0.267}$\pm$0.017 & $+$6.7  & 0.200 & \textbf{0.210}$\pm$0.017 & $+$1.0 \\
    HVAC   & 0.188 & 0.188$\pm$0.055 & \textbf{0.326}$\pm$0.012 & $+$13.9 & 0.146 & \textbf{0.299}$\pm$0.032 & $+$15.3 \\
    RE1-OB & 0.784 & 0.883$\pm$0.005 & \textbf{0.891}$\pm$0.012 & $+$10.7 & 0.784 & \textbf{0.896}$\pm$0.014 & $+$11.2 \\
    RE1-SS & 0.856 & 0.859$\pm$0.009 & \textbf{0.928}$\pm$0.000 & $+$7.2  & 0.856 & \textbf{0.877}$\pm$0.005 & $+$2.1 \\
    RE1-TT & 0.560 & 0.688$\pm$0.024 & \textbf{0.741}$\pm$0.009 & $+$18.1 & 0.560 & \textbf{0.795}$\pm$0.020 & $+$23.5 \\
    \midrule
    Mean   & 0.483 & 0.543 & \textbf{0.598} & $+$11.6 & 0.463 & \textbf{0.577} & $+$11.4 \\
    \bottomrule
  \end{tabular}
\end{table}

\paragraph{Domain knowledge under controlled retrieval.}
With retrieval guaranteed, DK raises top@1 on \emph{every} benchmark
(Table~\ref{tab:recall-controlled}): clearly on HVAC, RE1-SS, and RE1-TT;
modestly on SWaT, where it hurt end to end; within noise on WADI and
RE1-OB. On SWaT, DK's benefit
concentrates on \emph{quiet-evidence} causes that statistics cannot
rank: on the 11 SWaT scenarios whose cause the retriever misses (33
scenario-runs), it is ranked first $0/33$ times without DK and $6/33$ with. These are exactly the causes retrieval misses, so under
the end-to-end protocol they score zero regardless of DK, leaving
visible only DK's prior-conflict cost on loud-evidence
scenarios ($24/72 \to 22/72$ scenario-runs). A component's value can thus be
invisible, or even appear negative, under top@$k$ alone.

\paragraph{Robustness.}
A second backbone (Llama-3.3-70B) still beats every baseline on
microservices, with CPS much harder (below the best baseline on SWaT); WADI top@1 stays at or above the
best same-pool baseline at every $T \in \{0,1,2\}$; de-identifying all data
and DK documents shifts top@1 by a few points in either direction
(Appendices~\ref{app:robustness}--\ref{app:tdet}); and perturbing
$t_{\mathrm{det}}$ by $\pm 1$--$2$ minutes is benign on microservices, while $\pm 5$ minutes costs SWaT
retrieval 17--22~pp.

\section{Conclusion} \label{sec:conclusion}

We decomposed RCA evaluation into retrieval and reranking and used it to audit four benchmarks. On industrial CPS benchmarks, deviation magnitude alone leaves a substantial fraction of true causes outside a $K{=}15$ candidate set, and graph-based methods never clearly beat the best statistical baseline on per-scenario, retrieved-pool, or long-window graphs. A multi-signal retriever narrows this retrieval gap, and a single-call LLM reranker over per-candidate evidence matches or improves top@1 accuracy on both kinds of benchmark under one fixed configuration, without a learned causal graph or labeled traces. Because the detector determines the retrieval substrate, anomaly detection and RCA must be co-designed. Controlling retrieval also showed that top@$k$ can hide a component's benefit, as it did for domain knowledge. Nevertheless, these findings come with limitations: the industrial CPS benchmarks are small and cover only water and building systems, multi-cause attacks count as solved when any one cause is ranked, the SWaT retrieval gains assume a detection timestamp accurate to better than 5 minutes, the retriever's settings were fixed on the same benchmarks, the domain-knowledge document is not yet ingested adaptively, and the LLM reranker is nondeterministic. In deployment, faster diagnosis can shorten incidents in safety-critical infrastructure such as water treatment and building management, but automation bias and miscalibrated reranker confidence on out-of-distribution faults are real risks, so we recommend presenting a short ranked list of candidates rather than a single answer, surfacing the per-candidate evidence behind each rank, and treating human-in-the-loop oversight as essential.\looseness=-1

\begin{ack}
This research is supported by Baker Hughes, the ARC Training Centre for Whole
Life Design of Carbon Neutral Infrastructure (IC230100015), and the ARC Centre
of Excellence for Automated Decision-Making and Society (CE200100005).
The authors declare no other competing
interests.
\end{ack}

\bibliography{ref}
\bibliographystyle{unsrtnat}

\newpage
\appendix

\FloatBarrier
\section{Reproducibility}
\label{app:reproducibility}

\paragraph{Code.}
The pipeline, baselines, prompts, domain-knowledge documents, and the
scripts behind every table are released with the paper (link in the
abstract); the README lists the command for each table. Per-dataset
settings are declared in one registry (\texttt{method/runners/\_datasets.py}).
This appendix describes the setup; edge cases are handled as in the
released code.

\paragraph{Retrieval signals.}
Let $\mathbf{X}_B$ and $\mathbf{X}_F$ be the baseline and fault windows split
at the detection timestamp $t_{\mathrm{det}}$. For each sensor $s_i$:

\begin{itemize}[leftmargin=*,nosep]
  \item $\phi_{\text{mag}}(s_i) = \max_{t \in F}\, z^{\text{rob}}_i(t)$,
  the maximum signed RobustScaler $z$-score in the fault window, with the scaler
  fit on $\mathbf{X}_B[s_i]$.
  \item $\phi_{\text{ons}}(s_i) = -t^{\star}_i$, where
  $t^{\star}_i = \min\{t \in F : |z^{\mu,\sigma}_i(t)| \geq 1.5\}$ is the first
  fault-window timestamp at which the value, standardized with the baseline
  mean and standard deviation, crosses $1.5$. Sensors that never cross are
  dropped. Ties at equal $t^{\star}_i$ keep the data's column order
  (alphabetical on HVAC); on HVAC, where most sensors first cross at the
  first fault-window sample, this ordering accounts for the onset lift in
  Table~\ref{tab:retrieval-recall}.
  \item $\phi_{\text{stc}}(s_i) = \mathbf{1}[\,s_i\,\text{discrete}\,] \cdot
  \mathbf{1}[\,\text{mode}(\mathbf{X}_F[s_i]) \neq \text{mode}(\mathbf{X}_B[s_i])\,]$,
  ranked by the first row that departs from the baseline mode. A sensor is
  discrete if it takes at most 5 unique values in either window.
\end{itemize}

The candidate set $\mathcal{C}$ at budget $K$ splits $K$ evenly across the
active scorers ($5{+}5{+}5$ at $K{=}15$), takes each scorer's top entries,
and deduplicates the union, so $|\mathcal{C}| \leq K$. Candidates are
ordered magnitude, then onset, then state-change. The $\mathbf{+ons}$ and
$\mathbf{+stc}$ columns of Table~\ref{tab:retrieval-recall} activate the
magnitude and onset scorers, and all three scorers, respectively.

\paragraph{Evidence records.}
Each candidate is shown to the LLM as one line with its rank, name,
$\phi_{\text{mag}}$, onset offset relative to $t_{\mathrm{det}}$, baseline
and fault-window means, and absolute and relative shift. The line does
not say which scorer selected the candidate:

\begin{verbatim}
  3. carts_cpu        |  z-score=4.2  |  +2s  |
       before=0.080 -> after=0.310 (Delta=+0.230, +287.5%)
\end{verbatim}

\paragraph{Reranking prompt.}
The no-DK and with-DK configurations differ only in the system prompt.

\textit{No-DK (\texttt{level=none}):}
\begin{quote}\small\ttfamily
You are an expert in root cause analysis of complex systems based on
time-series anomaly evidence.
\end{quote}

\textit{With-DK (\texttt{level=light}):}
\begin{quote}\small\ttfamily
You are an expert in \{domain\} and root cause analysis. Use the
following operational system documentation to inform your reasoning:\\[2pt]
\{$\mathcal{D}$\}
\end{quote}

\noindent Here \texttt{\{domain\}} is a short dataset phrase (e.g., ``water
distribution ICS systems'' for WADI) and \texttt{\{$\mathcal{D}$\}} is the
dataset's domain-knowledge document (\texttt{method/prompts/}). The user
prompt is the same for all datasets:

\begin{quote}\small\ttfamily
A fault has been detected. The following items are candidates for root
cause analysis, ranked by deviation from baseline behaviour. Each line
shows the item name, its deviation magnitude, when it first deviated, and
its before/after values.\\[2pt]
\{anomaly\_summary\}\\[2pt]
Respond with a single JSON object containing BOTH a reasoning trace\\
and the ranked list:\\[2pt]
\quad\{\\
\quad\quad "reasoning": "<your detailed step-by-step analysis identifying the root cause and explaining why each top-ranked item was chosen over others>",\\
\quad\quad "ranked": ["item1", "item2", ...]\\
\quad\}\\[2pt]
Only include items from the provided list. Output only the JSON.
\end{quote}
\noindent Candidates appear in retrieval order, despite the phrase ``ranked
by deviation''. The prompt prescribes no reasoning steps or domain
heuristics, and the \texttt{reasoning} field is not scored. The prompt was
fixed before evaluation and was not iterated against test outcomes.

\paragraph{Decoding.}
The reranker is \texttt{openai/gpt-oss-120b} served through Groq's
OpenAI-compatible API, sampled at $T{=}1.0$ with
\texttt{max\_tokens}${=}4096$ and no seed, with one call per scenario and
$n{=}3$ runs per configuration. From the returned \texttt{ranked} list we
discard names outside $\mathcal{C}$ and append omitted candidates in
retrieval order.

\paragraph{Domain-knowledge documents.}
Each document $\mathcal{D}$ is drafted from the public system documentation
and describes component roles, propagation pathways, control loops, and
naming conventions, with general operational heuristics of the kind such
documentation carries. It never names a scenario's root cause or includes
fault labels, and it was not tuned on test outcomes. The documents are
released so readers can inspect them.

\FloatBarrier
\section{Datasets}
\label{app:datasets}

Each scenario has a baseline window, a fault window, a detection timestamp
$t_{\mathrm{det}}$, and its ground-truth root-cause metric(s).
$t_{\mathrm{det}}$ is the labeled attack start (WADI, SWaT), the injection
time (RCAEval), or the first occupied row of the fault day (HVAC). On
WADI, SWaT, and RCAEval this sets $t_{\mathrm{det}} = t_{\mathrm{fault}}$, as if
the system were detected without delay: the best case for any detector, in which
only the per-sensor lags $\delta_i$ remain. Every
method sees the same windows.

\paragraph{Unobservable causes.}
In two scenarios the ground-truth sensor is outside the evaluated set: WADI
attack~13 targets the setpoint \texttt{2\_PIC\_003\_SP}, which our sensor
filter removes, and SWaT attack~4 targets \texttt{MV504}, which is not
recorded. Both stay in every end-to-end evaluation as misses for all
methods, so scores remain comparable with published results. Only the
retrieval-controlled analysis (Section~\ref{sec:controlled}) excludes them,
since no pool spot can be reserved for an unobserved sensor.

\paragraph{WADI ($n{=}14$) and SWaT ($n{=}36$).}
Both are sampled at 1\,Hz, and each scenario uses a 30-minute baseline
immediately before the attack plus the attack window. On WADI we keep
seven sensor types (\texttt{MV}, \texttt{LS}, \texttt{LT}, \texttt{FIT},
\texttt{AIT}, \texttt{MCV}, \texttt{P}) following
LEMMA-RCA~\cite{zheng2024lemma} and merge the redundant
\texttt{2A\_*}/\texttt{2B\_*} sensors. Ground truth is the set of
manipulated sensors listed by the dataset authors. SWaT actuators
(\texttt{MV}, \texttt{P}, \texttt{UV}) are discrete and therefore eligible
for $\phi_{\text{stc}}$; the baselines treat them as continuous.

\paragraph{HVAC ($n{=}48$).}
Fault days from the LBNL Fault Detection and Diagnostics dataset (ORNL
Experimental RTU)~\cite{OEDI_Dataset_5763}: three fault types
(\texttt{SA\_temp\_bias}, \texttt{OA\_damper\_stuck},
\texttt{Inc\_Eco\_SP}), four seasons, and four fault levels, sampled every
minute. The baseline is the last full occupied day of the matching
fault-free file, a different day from the fault (so the two windows also
differ in weather), followed by the occupied
rows of the fault day. Each fault is active from 00:00 of the fault day,
so $t_{\mathrm{det}}$ (07:00) falls about 7\,h after onset. We
restrict to the occupied schedule (7:00--22:00) given in the LBNL
documentation, because the air-handling unit parks the outdoor-air damper
and several sensors stop reading outside it. Ground truth is
\texttt{RTU\_SA\_TEMP} for the supply-air bias and
\texttt{RTU\_OA\_DMPR\_DM} for the damper and economizer faults.

\paragraph{RCAEval RE1 ($n{=}125$ per suite).}
Online Boutique (RE1-OB), Sock Shop (RE1-SS), and Train Ticket (RE1-TT),
sampled at 1\,s, with baseline and fault windows of about 6 minutes on
each side of the injection, as in RCAEval~\cite{pham2025rcaeval}.
Preprocessing and service-level scoring follow RCAEval's published
protocol, which reproduces its reported BARO numbers.

\FloatBarrier
\section{Baselines}
\label{app:baselines}

All baselines use the same windows as the LLM pipeline, with the
implementations and defaults of RCAEval~\cite{pham2025rcaeval} and the
upstream libraries.

\paragraph{Statistical baselines.}
\textbf{BARO}~\cite{pham2024baro} runs through RCAEval's own entry point on
RCAEval and as a local maximum-absolute RobustScaler score on the CPS
datasets. \textbf{RCD}~\cite{ikram2022root} (\texttt{top\_k}${=}10$,
$\alpha_{\text{limit}}{=}0.5$) and
\textbf{$\epsilon$-Diagnosis}~\cite{shan2019diagnosis}
(\texttt{root\_cause\_top\_k}${=}10$) use PyRCA with patch sizes of 60
(WADI/SWaT), 4 (HVAC), and 100 (RCAEval). Both are seeded on HVAC and are
single unseeded runs elsewhere.

\paragraph{Graph-based baselines.}
PC and FCI~\cite{spirtes2000causation} (\texttt{causal-learn},
$\alpha{=}0.05$, Fisher-$z$) learn one graph per scenario from up to
10 minutes of data on each side of $t_{\mathrm{det}}$ on WADI/SWaT,
5 minutes on RCAEval, and 120 minutes on HVAC, after constant,
near-constant, and near-duplicate columns are removed. Undirected edges become two directed
edges. Three scorers run on each graph:
\textbf{CIRCA}~\cite{li2022causal} (upstream implementation,
\texttt{RHTScorer} followed by \texttt{DAScorer}, fit on the baseline rows
and tested on the fault rows);
\textbf{PageRank}~\cite{Page1998PageRank} (\texttt{scikit-network}) on the
transposed graph, so walks move toward upstream causes; and
\textbf{RandomWalk}~\cite{tong2006fast} on the edge-inverted graph (seed
42). PageRank and RandomWalk rank only the sensors in the graph, CIRCA
ranks all sensors, and a cause missing from a ranking counts as a miss.
CIRCA breaks ties by set order, so all runs fix
\texttt{PYTHONHASHSEED}${=}0$.

\FloatBarrier
\section{Asset Licenses and Versions}
\label{app:licenses}

\paragraph{Datasets.}
WADI~\cite{ahmed2017wadi} and SWaT~\cite{goh2016dataset} are distributed by
iTrust (Singapore University of Technology and Design) on request for
research use and are used under those terms. The HVAC data are the LBNL
Fault Detection and Diagnostics Datasets~\cite{OEDI_Dataset_5763}, obtained
through the Open Energy Data Initiative (DOI \texttt{10.25984/1881324})
under CC-BY-4.0. RCAEval~\cite{pham2025rcaeval} data and code are used
under its MIT license.

\paragraph{Code.}
Baselines come from RCAEval (MIT), PyRCA (BSD-3-Clause) for RCD and
$\epsilon$-Diagnosis, \texttt{causal-learn} (MIT) for PC/FCI,
\texttt{scikit-network} (BSD-3-Clause) for PageRank, and the NetManAIOps
CIRCA implementation under its repository license. Dependencies are pinned
in \texttt{requirements.txt} and \texttt{requirements-lock.txt}.

\paragraph{Models.}
The reranker backbones are \texttt{openai/gpt-oss-120b} (Apache-2.0) and
\texttt{llama-3.3-70b-versatile} (Llama~3.3 Community License), both served
through Groq's API under its terms of service. The domain-knowledge
documents were drafted, and rewritten for the de-identification control,
with Anthropic's Claude under its terms of service.

\FloatBarrier
\section{Full Result Grids and Saturation Curves}
\label{app:full-grids}

\begin{figure}[t]
  \centering
  \includegraphics[width=\linewidth]{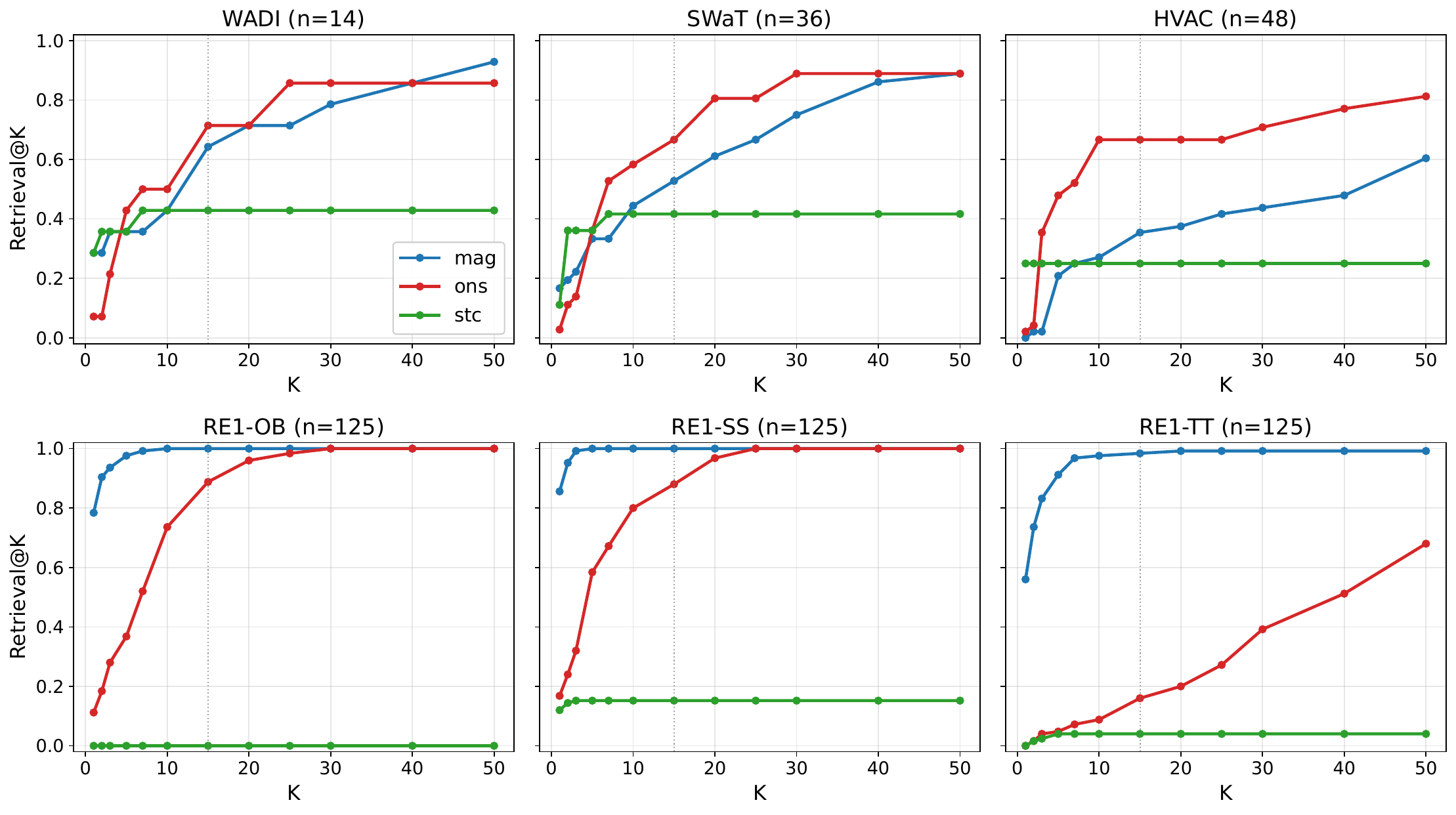}
  \caption{Retrieval saturation per signal. Each signal is allocated
  its own full $K$. Grey dotted: $K{=}15$ reference. The panels show
  three distinct saturation regimes (Amplitude-suppressed: HVAC;
  Rank-displaced: WADI, SWaT;
  Magnitude-dominant: RCAEval). Ties at equal onset
  time follow column order, which shapes the HVAC and WADI onset curves.}
  \label{fig:saturation}
\end{figure}

\begin{table}[t]
  \caption{Full end-to-end grid, industrial CPS benchmarks
  (expands Table~\ref{tab:main-results}). LLM rows: mean$\pm$std
  over $n{=}3$ runs at $T{=}1.0$; other rows single runs (RCD and
  $\epsilon$-Diagnosis seeded on HVAC, unseeded elsewhere). HVAC LLM rows
  are scored as in Appendix~\ref{app:hvac-correction}.}
  \label{tab:main-results-cps}
  \centering
  \resizebox{\textwidth}{!}{%
  \begin{tabular}{l cccc cccc cccc}
    \toprule
    & \multicolumn{4}{c}{WADI ($n{=}14$)} & \multicolumn{4}{c}{SWaT ($n{=}36$)} & \multicolumn{4}{c}{HVAC ($n{=}48$)} \\
    \cmidrule(lr){2-5}\cmidrule(lr){6-9}\cmidrule(lr){10-13}
    Method & top@1 & top@3 & top@5 & Avg@5 & top@1 & top@3 & top@5 & Avg@5 & top@1 & top@3 & top@5 & Avg@5 \\
    \midrule
    \multicolumn{13}{l}{\emph{Statistical baselines}} \\
    BARO                    & 0.214 & 0.357 & 0.357 & 0.314 & 0.194 & 0.306 & 0.417 & 0.306 & 0.000 & 0.021 & 0.062 & 0.025 \\
    RCD                     & 0.071 & 0.214 & 0.214 & 0.186 & 0.083 & 0.111 & 0.111 & 0.106 & 0.083 & 0.146 & 0.146 & 0.133 \\
    $\epsilon$-Diagnosis    & 0.000 & 0.071 & 0.071 & 0.043 & 0.028 & 0.056 & 0.083 & 0.061 & 0.146 & 0.146 & 0.146 & 0.146 \\
    \midrule
    \multicolumn{13}{l}{\emph{Graph-based baselines (per-scenario PC/FCI graphs)}} \\
    CIRCA (PC)              & 0.214 & 0.286 & 0.429 & 0.329 & 0.083 & 0.222 & 0.333 & 0.222 & 0.062 & 0.125 & 0.125 & 0.100 \\
    CIRCA (FCI)             & 0.214 & 0.357 & 0.429 & 0.343 & 0.111 & 0.222 & 0.333 & 0.228 & 0.104 & 0.125 & 0.125 & 0.117 \\
    PageRank (PC)           & 0.000 & 0.071 & 0.214 & 0.071 & 0.028 & 0.139 & 0.194 & 0.117 & 0.021 & 0.042 & 0.062 & 0.046 \\
    PageRank (FCI)          & 0.071 & 0.071 & 0.214 & 0.114 & 0.000 & 0.028 & 0.139 & 0.044 & 0.000 & 0.062 & 0.062 & 0.050 \\
    RandomWalk (PC)         & 0.071 & 0.143 & 0.286 & 0.186 & 0.000 & 0.167 & 0.222 & 0.128 & 0.000 & 0.021 & 0.021 & 0.013 \\
    RandomWalk (FCI)        & 0.000 & 0.071 & 0.143 & 0.071 & 0.000 & 0.056 & 0.056 & 0.044 & 0.000 & 0.000 & 0.021 & 0.008 \\
    \midrule
    \multicolumn{13}{l}{\emph{Multi-evidence LLM reranker}} \\
    LLM (no DK)
        & \textbf{0.333}$\pm$0.082 & 0.381$\pm$0.041 & 0.381$\pm$0.041 & 0.362$\pm$0.058
        & \textbf{0.213}$\pm$0.016 & \textbf{0.472}$\pm$0.000 & 0.481$\pm$0.016 & \textbf{0.402}$\pm$0.008
        & 0.153$\pm$0.043 & 0.243$\pm$0.032 & 0.292$\pm$0.021 & 0.239$\pm$0.031 \\
    LLM (with DK)
        & 0.310$\pm$0.041 & \textbf{0.476}$\pm$0.041 & \textbf{0.500}$\pm$0.000 & \textbf{0.424}$\pm$0.022
        & 0.148$\pm$0.016 & 0.380$\pm$0.032 & \textbf{0.509}$\pm$0.058 & 0.343$\pm$0.023
        & \textbf{0.278}$\pm$0.012 & \textbf{0.354}$\pm$0.021 & \textbf{0.382}$\pm$0.012 & \textbf{0.340}$\pm$0.009 \\
    \bottomrule
  \end{tabular}%
  }
\end{table}

\begin{table}[t]
  \caption{Full end-to-end grid, RCAEval microservice suites
  (expands Table~\ref{tab:main-results}). LLM rows report
  mean$\pm$std across $n{=}3$ independent runs at temperature
  $1.0$. Other rows are single runs (RCD and $\epsilon$-Diagnosis
  unseeded).}
  \label{tab:main-results-rcaeval}
  \centering
  \resizebox{\textwidth}{!}{%
  \begin{tabular}{l cccc cccc cccc}
    \toprule
    & \multicolumn{4}{c}{RE1-OB ($n{=}125$)} & \multicolumn{4}{c}{RE1-SS ($n{=}125$)} & \multicolumn{4}{c}{RE1-TT ($n{=}125$)} \\
    \cmidrule(lr){2-5}\cmidrule(lr){6-9}\cmidrule(lr){10-13}
    Method & top@1 & top@3 & top@5 & Avg@5 & top@1 & top@3 & top@5 & Avg@5 & top@1 & top@3 & top@5 & Avg@5 \\
    \midrule
    \multicolumn{13}{l}{\emph{Statistical baselines}} \\
    BARO                    & 0.784 & 0.936 & \textbf{0.976} & 0.912 & 0.856 & 0.992 & \textbf{1.000} & 0.962 & 0.560 & 0.856 & \textbf{0.952} & 0.810 \\
    RCD                     & 0.304 & 0.432 & 0.432 & 0.403 & 0.248 & 0.512 & 0.512 & 0.446 & 0.144 & 0.184 & 0.184 & 0.176 \\
    $\epsilon$-Diagnosis    & 0.072 & 0.224 & 0.336 & 0.221 & 0.232 & 0.496 & 0.616 & 0.459 & 0.008 & 0.072 & 0.112 & 0.062 \\
    \midrule
    \multicolumn{13}{l}{\emph{Graph-based baselines (per-scenario PC/FCI graphs)}} \\
    CIRCA (PC)              & 0.536 & 0.768 & 0.936 & 0.754 & 0.632 & 0.904 & 0.976 & 0.845 & 0.320 & 0.568 & 0.736 & 0.544 \\
    CIRCA (FCI)             & 0.576 & 0.816 & 0.920 & 0.778 & 0.624 & 0.864 & 0.976 & 0.829 & 0.328 & 0.512 & 0.624 & 0.496 \\
    PageRank (PC)           & 0.112 & 0.280 & 0.448 & 0.278 & 0.136 & 0.376 & 0.592 & 0.365 & 0.008 & 0.056 & 0.168 & 0.075 \\
    PageRank (FCI)          & 0.096 & 0.320 & 0.552 & 0.334 & 0.184 & 0.488 & 0.672 & 0.453 & 0.048 & 0.112 & 0.184 & 0.118 \\
    RandomWalk (PC)         & 0.080 & 0.328 & 0.600 & 0.338 & 0.112 & 0.416 & 0.656 & 0.398 & 0.064 & 0.128 & 0.160 & 0.122 \\
    RandomWalk (FCI)        & 0.112 & 0.352 & 0.552 & 0.331 & 0.144 & 0.368 & 0.528 & 0.358 & 0.072 & 0.144 & 0.168 & 0.131 \\
    \midrule
    \multicolumn{13}{l}{\emph{Multi-evidence LLM reranker}} \\
    LLM (no DK)
        & 0.875$\pm$0.009 & 0.952$\pm$0.000 & 0.968$\pm$0.000 & 0.937$\pm$0.004
        & 0.872$\pm$0.024 & 0.989$\pm$0.009 & \textbf{1.000}$\pm$0.000 & 0.964$\pm$0.009
        & 0.653$\pm$0.024 & 0.821$\pm$0.005 & 0.912$\pm$0.000 & 0.809$\pm$0.006 \\
    LLM (with DK)
        & \textbf{0.888}$\pm$0.008 & \textbf{0.960}$\pm$0.000 & \textbf{0.976}$\pm$0.000 & \textbf{0.946}$\pm$0.003
        & \textbf{0.944}$\pm$0.008 & \textbf{0.995}$\pm$0.005 & \textbf{1.000}$\pm$0.000 & \textbf{0.979}$\pm$0.003
        & \textbf{0.685}$\pm$0.009 & \textbf{0.864}$\pm$0.014 & 0.907$\pm$0.009 & \textbf{0.834}$\pm$0.013 \\
    \bottomrule
  \end{tabular}%
  }
\end{table}

\FloatBarrier
\section{HVAC LLM Scoring}
\label{app:hvac-correction}

HVAC's LLM runs use the retrieval-controlled pools of
Section~\ref{sec:controlled}: the retriever's $K{=}15$ pool with a spot
reserved for the true cause, which the retriever finds in 30 of 48
scenarios and which is inserted in the other 18. The end-to-end HVAC rows
of Table~\ref{tab:main-results} score the same predictions and count those
18 scenarios as misses. This matches the end-to-end protocol, because the
other 30 pools are the same with or without the reserved spot. The HVAC
baseline window is described in Appendix~\ref{app:datasets}.

\FloatBarrier
\section{Controlled Comparisons}
\label{app:controls}

\paragraph{Same candidates.}
Every control ranks exactly the $K{=}15$ candidate list the LLM was shown,
checked against the prompts sent to the LLM, with the same scoring as the
main tables. As a check, on WADI, SWaT, and RCAEval, Retrieval@15 of these lists
reproduces the +stc column of Table~\ref{tab:retrieval-recall}, and ordering them by
$\phi_{\text{mag}}$ reproduces BARO's published top@1 on all three
RCAEval suites. The fusion control is an equal-weight Borda count over the
magnitude, onset, and state-change ranks. Tables~\ref{tab:same-candidate-cps}
and~\ref{tab:same-candidate-ms} give the full grids, and
Table~\ref{tab:same-candidate-hvac} gives HVAC on its retriever pools
(Appendix~\ref{app:hvac-correction}).

\begin{table}[t]
  \caption{Same-candidate controls, industrial CPS (the candidate lists
  the LLM was shown; top@1 / top@3 / top@5 / Avg@5). LLM rows are the
  $n{=}3$ runs on the identical lists.}
  \label{tab:same-candidate-cps}
  \centering
  \footnotesize
  \setlength{\tabcolsep}{5pt}
  \begin{tabular}{lcc}
    \toprule
    Ranker (same candidate set $\mathcal{C}$) & WADI & SWaT \\
    \midrule
    mag-on-$\mathcal{C}$ ($\approx$BARO) & 0.286 / 0.357 / 0.357 / 0.329 & 0.167 / 0.222 / 0.333 / 0.244 \\
    ons-on-$\mathcal{C}$                & 0.214 / 0.286 / 0.500 / 0.329 & 0.194 / 0.278 / 0.417 / 0.311 \\
    stc-on-$\mathcal{C}$                & 0.286 / 0.357 / 0.429 / 0.357 & 0.139 / 0.389 / 0.389 / 0.339 \\
    fusion-on-$\mathcal{C}$ (Borda)     & 0.286 / \textbf{0.500} / \textbf{0.571} / \textbf{0.443} & 0.194 / 0.306 / 0.361 / 0.294 \\
    LLM (no DK, $n{=}3$)                & \textbf{0.333} / 0.381 / 0.381 / 0.362 & \textbf{0.213} / \textbf{0.472} / \textbf{0.481} / \textbf{0.402} \\
    \bottomrule
  \end{tabular}
\end{table}

\begin{table}[t]
  \caption{Same-candidate controls, RCAEval microservices
  (service-level; top@1 / top@3 / top@5 / Avg@5).}
  \label{tab:same-candidate-ms}
  \centering
  \footnotesize
  \setlength{\tabcolsep}{4pt}
  \resizebox{\textwidth}{!}{%
  \begin{tabular}{lccc}
    \toprule
    Ranker (same candidate set $\mathcal{C}$) & RE1-OB & RE1-SS & RE1-TT \\
    \midrule
    mag-on-$\mathcal{C}$ ($\approx$BARO) & 0.784 / 0.936 / \textbf{0.976} / 0.912 & 0.856 / \textbf{0.992} / \textbf{1.000} / 0.962 & 0.560 / \textbf{0.856} / \textbf{0.912} / 0.802 \\
    ons-on-$\mathcal{C}$                & 0.240 / 0.336 / 0.696 / 0.403 & 0.368 / 0.576 / 0.864 / 0.594 & 0.176 / 0.224 / 0.232 / 0.214 \\
    stc-on-$\mathcal{C}$                & 0.552 / 0.920 / \textbf{0.976} / 0.837 & 0.360 / 0.968 / \textbf{1.000} / 0.810 & 0.000 / 0.128 / 0.360 / 0.157 \\
    fusion-on-$\mathcal{C}$ (Borda)     & 0.464 / 0.912 / 0.960 / 0.818 & 0.512 / 0.936 / 0.992 / 0.846 & 0.224 / 0.592 / 0.760 / 0.542 \\
    LLM (no DK, $n{=}3$)                & \textbf{0.875} / \textbf{0.952} / 0.968 / \textbf{0.937} & \textbf{0.872} / 0.989 / \textbf{1.000} / \textbf{0.964} & \textbf{0.653} / 0.821 / \textbf{0.912} / \textbf{0.809} \\
    \bottomrule
  \end{tabular}%
  }
\end{table}

\begin{table}[t]
  \caption{Same-candidate controls on HVAC
  (retrieval-controlled retriever pools, top@1). State-change beats the
  no-DK LLM here but not the with-DK LLM, and it scores 0.000 top@1 on
  RE1-TT (Table~\ref{tab:same-candidate-ms}).}
  \label{tab:same-candidate-hvac}
  \centering
  \footnotesize
  \begin{tabular}{lc}
    \toprule
    Ranker & top@1 \\
    \midrule
    mag-on-$\mathcal{C}$    & 0.000 \\
    ons-on-$\mathcal{C}$    & 0.104 \\
    stc-on-$\mathcal{C}$    & 0.250 \\
    fusion-on-$\mathcal{C}$ (Borda) & 0.104 \\
    LLM (no DK, $n{=}3$)    & 0.188$\pm$0.055 \\
    LLM (with DK, $n{=}3$)  & \textbf{0.326}$\pm$0.012 \\
    \bottomrule
  \end{tabular}
\end{table}

\paragraph{Retrieval-controlled pools.}
Retrieval@$K$ is 1 by construction. The retriever pool is the $K{=}15$
multi-signal union, with the true cause inserted when the retriever misses
it, and the all-candidates pool contains every sensor. All nine baselines
are re-run on the same pools; graph methods fit their graphs on the
pool-restricted data, and stochastic baselines are seeded (except RCD
on the RCAEval all-candidates pools, which reuses its end-to-end run). Because RCD and
$\epsilon$-Diagnosis are unseeded in the end-to-end tables, their cells can
differ between the two protocols; no best-baseline value in
Table~\ref{tab:recall-controlled} depends on an unseeded run. The two
unobservable-cause scenarios (Appendix~\ref{app:datasets}) are excluded, so
WADI and SWaT have 13 and 35 scenarios here.
Tables~\ref{tab:pool-config1} and~\ref{tab:pool-config2} give the full grids.

An inserted cause replaces the last candidate at a seeded random position,
while the other candidates keep their order. Baselines ignore this order,
but the LLM does not: on HVAC the inserted cause is ranked first in 5/27
(no-DK) and 7/27 (with-DK) scenario-runs when it lands in the top five
slots, and in 0/27 otherwise. Counting such hits as misses lowers HVAC
no-DK from $0.188$ to $0.153$ (below RCD's $0.188$), HVAC with-DK from
$0.326$ to $0.278$, RE1-TT from $0.688/0.741$ to $0.675/0.725$, and SWaT
with-DK from $0.267$ to $0.248$. Every with-DK lead in
Table~\ref{tab:recall-controlled} remains.

We also ran a stress test with random pools ($K{-}1$ randomly drawn
non-cause sensors plus the cause). There a baseline beats the LLM in
three cells: RE1-TT (BARO $0.912$ vs.\ $0.885$), SWaT (CIRCA-FCI $0.400$
vs.\ $0.362$, the only setup where a graph method beats the best
statistical baseline), and RE1-SS (BARO $0.928$ vs.\ $0.744$). On
RCAEval, the random pool keeps only one metric of the faulty service,
which leaves the LLM little corroborating evidence.

\begin{table}[t]
  \caption{Retrieval-controlled grid, retriever pool ($K{=}15$,
  reserved ground-truth spot): top@1 for every baseline on the
  identical pools. LLM row: no-DK, $n{=}3$.}
  \label{tab:pool-config1}
  \centering
  \footnotesize
  \setlength{\tabcolsep}{5pt}
  \begin{tabular}{lcccccc}
    \toprule
    Method & WADI & SWaT & RE1-OB & RE1-SS & RE1-TT & HVAC \\
    \midrule
    BARO                 & 0.308 & 0.200 & 0.784 & 0.856 & 0.560 & 0.000 \\
    RCD                  & 0.154 & 0.114 & 0.440 & 0.384 & 0.280 & \textbf{0.188} \\
    $\epsilon$-Diagnosis & 0.077 & 0.057 & 0.176 & 0.368 & 0.104 & 0.146 \\
    PageRank (PC)        & 0.077 & 0.086 & 0.304 & 0.472 & 0.192 & 0.062 \\
    PageRank (FCI)       & 0.154 & 0.114 & 0.296 & 0.352 & 0.176 & 0.083 \\
    RandomWalk (PC)      & 0.077 & 0.114 & 0.264 & 0.480 & 0.208 & 0.042 \\
    RandomWalk (FCI)     & 0.231 & 0.114 & 0.216 & 0.392 & 0.224 & 0.021 \\
    CIRCA (PC)           & 0.231 & 0.200 & 0.704 & 0.760 & 0.480 & 0.083 \\
    CIRCA (FCI)          & 0.154 & 0.114 & 0.720 & 0.744 & 0.520 & 0.042 \\
    LLM (no DK, $n{=}3$) & \textbf{0.410} & \textbf{0.229} & \textbf{0.883} & \textbf{0.859} & \textbf{0.688} & \textbf{0.188} \\
    \bottomrule
  \end{tabular}
\end{table}

\begin{table}[t]
  \caption{Retrieval-controlled grid, all candidates (every sensor in
  the pool): top@1 for every baseline. LLM row: no-DK, $n{=}3$.}
  \label{tab:pool-config2}
  \centering
  \footnotesize
  \setlength{\tabcolsep}{5pt}
  \begin{tabular}{lcccccc}
    \toprule
    Method & WADI & SWaT & RE1-OB & RE1-SS & RE1-TT & HVAC \\
    \midrule
    BARO                 & 0.231 & 0.200 & 0.784 & 0.856 & 0.560 & 0.000 \\
    RCD                  & 0.231 & 0.086 & 0.304 & 0.248 & 0.144 & 0.083 \\
    $\epsilon$-Diagnosis & 0.000 & 0.029 & 0.072 & 0.232 & 0.008 & 0.146 \\
    PageRank (PC)        & 0.000 & 0.029 & 0.112 & 0.136 & 0.008 & 0.021 \\
    PageRank (FCI)       & 0.077 & 0.000 & 0.096 & 0.184 & 0.048 & 0.000 \\
    RandomWalk (PC)      & 0.077 & 0.000 & 0.080 & 0.112 & 0.064 & 0.000 \\
    RandomWalk (FCI)     & 0.000 & 0.000 & 0.112 & 0.144 & 0.072 & 0.000 \\
    CIRCA (PC)           & 0.231 & 0.086 & 0.536 & 0.632 & 0.320 & 0.062 \\
    CIRCA (FCI)          & 0.231 & 0.114 & 0.576 & 0.624 & 0.328 & 0.104 \\
    LLM (no DK, $n{=}3$) & \textbf{0.385} & \textbf{0.210} & \textbf{0.896} & \textbf{0.877} & \textbf{0.795} & \textbf{0.299} \\
    \bottomrule
  \end{tabular}
\end{table}

\FloatBarrier
\section{Robustness Controls: Backbone, Temperature, Memorization,
Cost}
\label{app:robustness}

\paragraph{Second backbone.}
We ran \texttt{llama-3.3-70b-versatile} with the same prompts on the same
retrieval-controlled retriever pools (Table~\ref{tab:backbone}). It beats
the best baseline on all three microservice suites (exceeding
gpt-oss-120b on RE1-SS), ties it on WADI, and falls below it on SWaT. It was
not run on HVAC.

\begin{table}[t]
  \caption{Backbone control: top@1 on identical retrieval-controlled
  retriever pools (no-DK, $n{=}3$).}
  \label{tab:backbone}
  \centering
  \footnotesize
  \begin{tabular}{lccc}
    \toprule
    Group & gpt-oss-120b & Llama-3.3-70B & Best baseline \\
    \midrule
    WADI   & \textbf{0.410}$\pm$0.044 & 0.308$\pm$0.000 & 0.308 (BARO) \\
    SWaT   & \textbf{0.229}$\pm$0.029 & 0.171$\pm$0.029 & 0.200 (BARO, CIRCA) \\
    RE1-OB & \textbf{0.883}$\pm$0.005 & 0.856$\pm$0.008 & 0.784 (BARO) \\
    RE1-SS & 0.859$\pm$0.009 & \textbf{0.893}$\pm$0.005 & 0.856 (BARO) \\
    RE1-TT & \textbf{0.688}$\pm$0.024 & 0.597$\pm$0.012 & 0.560 (BARO) \\
    \bottomrule
  \end{tabular}
\end{table}

\paragraph{Decoding temperature.}
On the WADI retriever pools we also ran $T{=}0$ and $T{=}2$, the provider
maximum (Table~\ref{tab:temperature}). Top@1 ranges from $0.308$ ($T{=}2$)
to $0.410$ ($T{=}1$) and never falls below the best same-pool baseline
($0.308$). Results vary across runs even at $T{=}0$, which is why every LLM
result is a mean over $n{=}3$ runs.

\begin{table}[t]
  \caption{Decoding-temperature control (WADI, retrieval-controlled
  retriever pool, no-DK, $n{=}3$ per temperature).}
  \label{tab:temperature}
  \centering
  \footnotesize
  \begin{tabular}{lccc}
    \toprule
    Temperature & top@1 & top@3 & top@5 \\
    \midrule
    $T{=}0$   & 0.359$\pm$0.044 & 0.538$\pm$0.000 & 0.590$\pm$0.044 \\
    $T{=}1.0$ (default) & 0.410$\pm$0.044 & 0.538$\pm$0.077 & 0.615$\pm$0.077 \\
    $T{=}2.0$ (provider max) & 0.308$\pm$0.000 & 0.513$\pm$0.044 & 0.590$\pm$0.044 \\
    \bottomrule
  \end{tabular}
\end{table}

\paragraph{De-identification.}
To test whether the reranker recalls the public benchmarks rather than
reading the evidence, we renamed every sensor with a seeded,
structure-preserving map (e.g., \texttt{1\_FIT\_001\_PV} $\to$
\texttt{S1\_FLOW\_01\_VALUE}), keeping sensor types, units, and values,
and rewrote the domain-knowledge documents in the renamed vocabulary with
all facility and application names removed. We then re-ran the
retriever-pool protocol (Table~\ref{tab:anonymization}). Top@1 changes by a
few points in either direction. The de-identified documents raise top@1 at
least as much as the real ones on four of six benchmarks, slightly less on
RE1-SS ($+5.6$ vs.\ $+6.9$~pp), and not at all on SWaT, suggesting that the
documents help mainly through the process structure they describe.

\begin{table}[t]
  \caption{De-identification control: top@1 on retriever pools
  ($n{=}3$ means) with real vs.\ anonymized sensor names and DK
  documents.}
  \label{tab:anonymization}
  \centering
  \footnotesize
  \setlength{\tabcolsep}{5pt}
  \begin{tabular}{lcccc}
    \toprule
    Group & real / no-DK & anon / no-DK & real / with-DK & anon / anon-DK \\
    \midrule
    WADI   & 0.410 & 0.410 & 0.436 & \textbf{0.487} \\
    SWaT   & 0.229 & \textbf{0.286} & 0.267 & \textbf{0.286} \\
    HVAC   & 0.188 & 0.139 & 0.326 & \textbf{0.375} \\
    RE1-OB & 0.883 & 0.848 & \textbf{0.891} & 0.880 \\
    RE1-SS & 0.859 & 0.877 & 0.928 & \textbf{0.933} \\
    RE1-TT & 0.688 & 0.645 & \textbf{0.741} & 0.731 \\
    \bottomrule
  \end{tabular}
\end{table}

\paragraph{Cost.}
In the runs that logged token usage (the WADI temperature runs), one
no-DK call at $K{=}15$ used about $0.8$k input and $1.2$k output tokens,
including reasoning tokens. Each scenario needs one call per run.

\FloatBarrier
\section{Detection-Timestamp Sensitivity}
\label{app:tdet}

We shifted $t_{\mathrm{det}}$, re-extracted the windows, and recomputed
Retrieval@15 of the candidate pool that combines all three signals,
magnitude, onset, and state change (Table~\ref{tab:tdet}). Scenarios whose
shifted fault window becomes too short are skipped, and on WADI/SWaT the
unshifted value is recomputed on the same subset, so the change reflects
the shift alone.

\begin{table}[t]
  \caption{Detection-timestamp sensitivity: change in Retrieval@15 of
  the three-signal candidate pool under $t_{\mathrm{det}}$ perturbation,
  matched scenario subsets on WADI/SWaT ($+5$ min skips short attacks and
  evaluates 8 of 14 WADI and 12 of 36 SWaT scenarios). RCAEval suites are perturbed at $\pm 1$
  and $\pm 2$ minutes (windows of about 6 minutes per side); the entry reports the
  maximum absolute change across the three suites. No HVAC
  scenarios are skipped at any offset.}
  \label{tab:tdet}
  \centering
  \footnotesize
  \setlength{\tabcolsep}{5pt}
  \begin{tabular}{lcc}
    \toprule
    Benchmark & $-5$ min & $+5$ min \\
    \midrule
    WADI & $0.0$ pp & $-12.5$ pp \\
    SWaT & $-22.2$ pp & $-16.7$ pp \\
    HVAC & $-27.1$ pp & $-12.5$ pp \\
    RCAEval $\times 3$ ($\pm 1$/$\pm 2$ min) & \multicolumn{2}{c}{$\leq 3.2$ pp at every tested offset} \\
    \bottomrule
  \end{tabular}
\end{table}

Microservice retrieval is robust to the shift, while SWaT retrieval is not.
On SWaT, an early $t_{\mathrm{det}}$ places routine actuator cycling at the
start of the fault window, where the onset and state-change scorers pick it
up; the HVAC and WADI changes depend on the order of tied onsets and vanish
when ties are broken by deviation size. A 5-minute error costs SWaT
retrieval 17--22~pp under either order, so
the SWaT gains depend on an accurate $t_{\mathrm{det}}$. This is why the
formulation in Sections~\ref{sec:motivation} and~\ref{sec:decomposition}
uses detected rather than true onsets
(Eq.~\eqref{eq:precedence}), and why detection and retrieval should be
designed together.

\FloatBarrier
\section{Graph-Based Methods Under Alternative Setups}
\label{app:graphs}

Expert-specified graphs do not exist for these benchmarks and are rarely
available in practice~\cite{ikram2022root}, so we test two alternatives to
per-scenario graphs.

\paragraph{Graphs on retrieval-controlled pools.}
Fitting per-scenario PC/FCI graphs on the ${\leq}15$-node pools, which
always contain the cause, does not close the gap on CPS. The best graph
method still trails the best baseline on WADI ($0.231$ vs.\ BARO's
$0.308$) and HVAC ($0.083$ vs.\ RCD's $0.188$) and only ties BARO on SWaT
($0.200$), while CIRCA improves on microservices (RE1-OB
$0.576 \to 0.720$; Tables~\ref{tab:pool-config1}--\ref{tab:pool-config2}).
The CPS failure is therefore not mainly a matter of graph size or pool
coverage.

\paragraph{Graphs from long normal-operation windows.}
We also learned one global PC/FCI graph per dataset from the multi-day
normal-operation data and ran the scorers on it
(Table~\ref{tab:longwindow}). On CPS this does not help: the best scorer
reaches $0.214$ on WADI (tying BARO), $0.083$ on SWaT (BARO $0.194$), and
$0.083$ on HVAC ($\epsilon$-Diagnosis $0.146$). On microservices the best
CIRCA variant improves but stays far below BARO (RE1-OB $0.576 \to 0.632$
vs.\ $0.784$; RE1-SS $0.632 \to 0.648$ vs.\ $0.856$). RE1-TT, the largest
suite, was not run, to bound compute. RCAEval has no continuous normal
data, so its graphs are learned from concatenated pre-injection baselines,
and HVAC's from its four concatenated seasonal fault-free files; the joins
may introduce spurious edges. For tractability, the WADI/SWaT data are
subsampled to 1-minute rows and the RCAEval data 1:5.

\begin{table}[t]
  \caption{Ranking heads on global PC/FCI graphs learned from
  multi-day normal-operation corpora (top@1).}
  \label{tab:longwindow}
  \centering
  \footnotesize
  \setlength{\tabcolsep}{5pt}
  \begin{tabular}{lccccc}
    \toprule
    Method & WADI & SWaT & HVAC & RE1-OB & RE1-SS \\
    \midrule
    PageRank (global PC)    & 0.071 & 0.000 & 0.000 & 0.104 & 0.184 \\
    RandomWalk (global PC)  & 0.000 & 0.000 & 0.000 & 0.176 & 0.208 \\
    CIRCA (global PC)       & 0.214 & 0.083 & 0.083 & 0.568 & 0.632 \\
    PageRank (global FCI)   & 0.071 & 0.000 & 0.000 & 0.000 & 0.216 \\
    RandomWalk (global FCI) & 0.000 & 0.000 & 0.000 & 0.000 & 0.192 \\
    CIRCA (global FCI)      & 0.214 & 0.028 & 0.083 & 0.632 & 0.648 \\
    \bottomrule
  \end{tabular}
\end{table}

\paragraph{Where the long-window graphs fail.}
Applying the decomposition, with retrieval failure meaning the cause is not
a node of the learned graph: on WADI the cause is in the graph in 71.4\% of
scenarios and on SWaT in 86.1\%, yet it is ranked first in at most 21.4\%
and 8.3\%. On HVAC, RE1-OB, and RE1-SS the cause is always in the graph, so
every failure is a reranking failure. We hypothesize that CPS feedback
loops, redundant sensors, and tightly coupled actuators are at odds with
DAG-based causal discovery at any sample size.

\FloatBarrier

\clearpage
\section*{NeurIPS Paper Checklist}

\begin{enumerate}

\item {\bf Claims}
    \item[] Question: Do the main claims made in the abstract and introduction accurately reflect the paper's contributions and scope?
    \item[] Answer: \answerYes{}. 
    \item[] Justification: The abstract and Section \ref{sec:intro} state the retrieval--reranking decomposition, the magnitude-only Retrieval@15 ceiling of 35--64\% on industrial CPS benchmarks (vs.\ 98--100\% on microservice benchmarks), that the single no-DK configuration matches or exceeds the best baseline's top@1 on all six suites (by up to $+$12), the retrieval-controlled comparisons (retriever and all-candidates pools) in which the LLM is at least as accurate as every baseline, and the $+$7 to $+$18 top@1 lead of the with-DK configuration over the best same-pool baseline under controlled retrieval; each number is reproduced by Tables~\ref{tab:retrieval-recall}--\ref{tab:recall-controlled} in Section~\ref{sec:results} and the full grids in Appendix~\ref{app:full-grids}.
    \item[] Guidelines:
    \begin{itemize}
        \item The answer \answerNA{} means that the abstract and introduction do not include the claims made in the paper.
        \item The abstract and/or introduction should clearly state the claims made, including the contributions made in the paper and important assumptions and limitations. A \answerNo{} or \answerNA{} answer to this question will not be perceived well by the reviewers. 
        \item The claims made should match theoretical and experimental results, and reflect how much the results can be expected to generalize to other settings. 
        \item It is fine to include aspirational goals as motivation as long as it is clear that these goals are not attained by the paper. 
    \end{itemize}

\item {\bf Limitations}
    \item[] Question: Does the paper discuss the limitations of the work performed by the authors?
    \item[] Answer: \answerYes{}.
    \item[] Justification: Section \ref{sec:conclusion} discusses the small per-dataset scenario counts on the industrial CPS benchmarks (WADI $n{=}14$, SWaT $n{=}36$, HVAC $n{=}48$), the limited domain coverage (two domains: industrial CPS, spanning water treatment, water distribution, and building HVAC, and microservices), the scoring of multi-cause attacks as solved when any one cause is ranked, the retrieval-contingent effect of the domain-knowledge document and its non-adaptive ingestion, the dependence of the SWaT retrieval gains on a detection timestamp accurate to better than 5 minutes (Appendix~\ref{app:tdet}), retriever settings fixed on the evaluated benchmarks, and the LLM reranker's nondeterminism (bounded by $n{=}3$ runs, a temperature sweep, a second backbone, and a de-identification control).
    \item[] Guidelines:
    \begin{itemize}
        \item The answer \answerNA{} means that the paper has no limitation while the answer \answerNo{} means that the paper has limitations, but those are not discussed in the paper. 
        \item The authors are encouraged to create a separate ``Limitations'' section in their paper.
        \item The paper should point out any strong assumptions and how robust the results are to violations of these assumptions (e.g., independence assumptions, noiseless settings, model well-specification, asymptotic approximations only holding locally). The authors should reflect on how these assumptions might be violated in practice and what the implications would be.
        \item The authors should reflect on the scope of the claims made, e.g., if the approach was only tested on a few datasets or with a few runs. In general, empirical results often depend on implicit assumptions, which should be articulated.
        \item The authors should reflect on the factors that influence the performance of the approach. For example, a facial recognition algorithm may perform poorly when image resolution is low or images are taken in low lighting. Or a speech-to-text system might not be used reliably to provide closed captions for online lectures because it fails to handle technical jargon.
        \item The authors should discuss the computational efficiency of the proposed algorithms and how they scale with dataset size.
        \item If applicable, the authors should discuss possible limitations of their approach to address problems of privacy and fairness.
        \item While the authors might fear that complete honesty about limitations might be used by reviewers as grounds for rejection, a worse outcome might be that reviewers discover limitations that aren't acknowledged in the paper. The authors should use their best judgment and recognize that individual actions in favor of transparency play an important role in developing norms that preserve the integrity of the community. Reviewers will be specifically instructed to not penalize honesty concerning limitations.
    \end{itemize}

\item {\bf Theory assumptions and proofs}
    \item[] Question: For each theoretical result, does the paper provide the full set of assumptions and a complete (and correct) proof?
    \item[] Answer: \answerNA{}.
    \item[] Justification: The paper does not contain theoretical results requiring proofs; the retrieval--reranking decomposition (Section \ref{sec:decomposition}) and the formal definitions in Eqs.~(1)--(4) are structural and definitional rather than theorem-bearing.
    \item[] Guidelines:
    \begin{itemize}
        \item The answer \answerNA{} means that the paper does not include theoretical results. 
        \item All the theorems, formulas, and proofs in the paper should be numbered and cross-referenced.
        \item All assumptions should be clearly stated or referenced in the statement of any theorems.
        \item The proofs can either appear in the main paper or the supplemental material, but if they appear in the supplemental material, the authors are encouraged to provide a short proof sketch to provide intuition. 
        \item Inversely, any informal proof provided in the core of the paper should be complemented by formal proofs provided in appendix or supplemental material.
        \item Theorems and Lemmas that the proof relies upon should be properly referenced. 
    \end{itemize}

    \item {\bf Experimental result reproducibility}
    \item[] Question: Does the paper fully disclose all the information needed to reproduce the main experimental results of the paper to the extent that it affects the main claims and/or conclusions of the paper (regardless of whether the code and data are provided or not)?
    \item[] Answer: \answerYes{}
    \item[] Justification: Section~\ref{sec:experiments} and Appendix~\ref{app:datasets} cover the benchmarks, window split, and $k/K$ values; Appendix~\ref{app:reproducibility} specifies the per-signal scorers ($\phi_{\text{mag}}$, $\phi_{\text{ons}}$, $\phi_{\text{stc}}$) with thresholds, the candidate-set construction and the budget allocation across signals, the per-candidate evidence schema, the system/user prompt templates with a worked example, decoding configuration ($T{=}1.0$, $n{=}3$ runs), and the domain-knowledge document $\mathcal{D}$ (the final document used per dataset is released alongside the code so readers can directly verify the absence of fault-label leakage).

    \item[] Guidelines:
    \begin{itemize}
        \item The answer \answerNA{} means that the paper does not include experiments.
        \item If the paper includes experiments, a \answerNo{} answer to this question will not be perceived well by the reviewers: Making the paper reproducible is important, regardless of whether the code and data are provided or not.
        \item If the contribution is a dataset and\slash or model, the authors should describe the steps taken to make their results reproducible or verifiable. 
        \item Depending on the contribution, reproducibility can be accomplished in various ways. For example, if the contribution is a novel architecture, describing the architecture fully might suffice, or if the contribution is a specific model and empirical evaluation, it may be necessary to either make it possible for others to replicate the model with the same dataset, or provide access to the model. In general. releasing code and data is often one good way to accomplish this, but reproducibility can also be provided via detailed instructions for how to replicate the results, access to a hosted model (e.g., in the case of a large language model), releasing of a model checkpoint, or other means that are appropriate to the research performed.
        \item While NeurIPS does not require releasing code, the conference does require all submissions to provide some reasonable avenue for reproducibility, which may depend on the nature of the contribution. For example
        \begin{enumerate}
            \item If the contribution is primarily a new algorithm, the paper should make it clear how to reproduce that algorithm.
            \item If the contribution is primarily a new model architecture, the paper should describe the architecture clearly and fully.
            \item If the contribution is a new model (e.g., a large language model), then there should either be a way to access this model for reproducing the results or a way to reproduce the model (e.g., with an open-source dataset or instructions for how to construct the dataset).
            \item We recognize that reproducibility may be tricky in some cases, in which case authors are welcome to describe the particular way they provide for reproducibility. In the case of closed-source models, it may be that access to the model is limited in some way (e.g., to registered users), but it should be possible for other researchers to have some path to reproducing or verifying the results.
        \end{enumerate}
    \end{itemize}

\item {\bf Open access to data and code}
    \item[] Question: Does the paper provide open access to the data and code, with sufficient instructions to faithfully reproduce the main experimental results, as described in supplemental material?
    \item[] Answer: \answerYes{}.
    \item[] Justification: All four datasets are existing public benchmarks (WADI, SWaT via iTrust request; HVAC/OEDI; RCAEval) and we cite the originals rather than re-host. Our retriever and LLM reranker pipeline, baselines, prompts, and scripts to reproduce every reported table are publicly released at \url{https://github.com/cruiseresearchgroup/DecompRCA}, accessible without a personal request. 
    \item[] Guidelines:
    \begin{itemize}
        \item The answer \answerNA{} means that paper does not include experiments requiring code.
        \item Please see the NeurIPS code and data submission guidelines (\url{https://neurips.cc/public/guides/CodeSubmissionPolicy}) for more details.
        \item While we encourage the release of code and data, we understand that this might not be possible, so \answerNo{} is an acceptable answer. Papers cannot be rejected simply for not including code, unless this is central to the contribution (e.g., for a new open-source benchmark).
        \item The instructions should contain the exact command and environment needed to run to reproduce the results. See the NeurIPS code and data submission guidelines (\url{https://neurips.cc/public/guides/CodeSubmissionPolicy}) for more details.
        \item The authors should provide instructions on data access and preparation, including how to access the raw data, preprocessed data, intermediate data, and generated data, etc.
        \item The authors should provide scripts to reproduce all experimental results for the new proposed method and baselines. If only a subset of experiments are reproducible, they should state which ones are omitted from the script and why.
        \item At submission time, to preserve anonymity, the authors should release anonymized versions (if applicable).
        \item Providing as much information as possible in supplemental material (appended to the paper) is recommended, but including URLs to data and code is permitted.
    \end{itemize}

\item {\bf Experimental setting/details}
    \item[] Question: Does the paper specify all the training and test details (e.g., data splits, hyperparameters, how they were chosen, type of optimizer) necessary to understand the results?
    \item[] Answer: \answerYes{}.
    \item[] Justification: Section~\ref{sec:experiments} specifies the per-dataset scenario counts and the evaluated $k$ and $K$ values, Appendix~\ref{app:datasets} the baseline-vs-fault window split, and Appendix~\ref{app:baselines} the baseline configurations (published defaults with dataset-specific patch sizes). No model training is performed since the pipeline is unsupervised and relies on a pretrained LLM. 
    \item[] Guidelines:
    \begin{itemize}
        \item The answer \answerNA{} means that the paper does not include experiments.
        \item The experimental setting should be presented in the core of the paper to a level of detail that is necessary to appreciate the results and make sense of them.
        \item The full details can be provided either with the code, in appendix, or as supplemental material.
    \end{itemize}

\item {\bf Experiment statistical significance}
    \item[] Question: Does the paper report error bars suitably and correctly defined or other appropriate information about the statistical significance of the experiments?    
    \item[] Answer: \answerYes{}.
        \item[] Justification: We run the LLM reranker $n{=}3$ times at $T{=}1.0$ per configuration and report the mean and the sample standard deviation over runs in Tables~\ref{tab:headroom}--\ref{tab:recall-controlled}, \ref{tab:main-results-cps}, and~\ref{tab:main-results-rcaeval}; a decoding-temperature sweep and a second backbone bound the remaining variability (Appendix~\ref{app:robustness}). The retriever, BARO, PageRank, RandomWalk (fixed seed), and, with \texttt{PYTHONHASHSEED}${=}0$, PC/FCI graph fitting and CIRCA are deterministic given the data; the stochastic baselines (RCD, $\epsilon$-Diagnosis) are single runs under the upstream defaults on WADI/SWaT/RCAEval and seeded on HVAC and in the controlled grids, except RCD on the RCAEval all-candidates pools, which reuses its end-to-end run (Appendix~\ref{app:controls}).
    \item[] Guidelines:
    \begin{itemize}
        \item The answer \answerNA{} means that the paper does not include experiments.
        \item The authors should answer \answerYes{} if the results are accompanied by error bars, confidence intervals, or statistical significance tests, at least for the experiments that support the main claims of the paper.
        \item The factors of variability that the error bars are capturing should be clearly stated (for example, train/test split, initialization, random drawing of some parameter, or overall run with given experimental conditions).
        \item The method for calculating the error bars should be explained (closed form formula, call to a library function, bootstrap, etc.)
        \item The assumptions made should be given (e.g., Normally distributed errors).
        \item It should be clear whether the error bar is the standard deviation or the standard error of the mean.
        \item It is OK to report 1-sigma error bars, but one should state it. The authors should preferably report a 2-sigma error bar than state that they have a 96\% CI, if the hypothesis of Normality of errors is not verified.
        \item For asymmetric distributions, the authors should be careful not to show in tables or figures symmetric error bars that would yield results that are out of range (e.g., negative error rates).
        \item If error bars are reported in tables or plots, the authors should explain in the text how they were calculated and reference the corresponding figures or tables in the text.
    \end{itemize}

\item {\bf Experiments compute resources}
    \item[] Question: For each experiment, does the paper provide sufficient information on the computer resources (type of compute workers, memory, time of execution) needed to reproduce the experiments?
    \item[] Answer: \answerYes{}
    \item[] Justification: The LLM reranker is served by Groq's inference API using \texttt{openai/gpt-oss-120b}; runs are stateless single-prompt completions, so the dominant cost is third-party API calls rather than local hardware. The optional domain-knowledge document $\mathcal{D}$ was drafted offline using Anthropic Claude over each dataset's public description, and is reused across all evaluation runs (it is not part of the per-scenario inference loop). The multi-signal retriever, statistical baselines (BARO, RCD, $\epsilon$-Diagnosis), and the per-scenario PC/FCI graph fits with CIRCA / PageRank / RandomWalk scorers all run on a single Apple Silicon MacBook (CPU only, no GPU) and complete in minutes per dataset, with two exceptions: the per-scenario FCI fits on RE1-TT, whose larger metric count makes each fit take about two minutes (a few hours for the suite), and the graphs of Appendix~\ref{app:graphs} learned from multi-day normal corpora (up to about six minutes per graph).
    \item[] Guidelines:
    \begin{itemize}
        \item The answer \answerNA{} means that the paper does not include experiments.
        \item The paper should indicate the type of compute workers CPU or GPU, internal cluster, or cloud provider, including relevant memory and storage.
        \item The paper should provide the amount of compute required for each of the individual experimental runs as well as estimate the total compute. 
        \item The paper should disclose whether the full research project required more compute than the experiments reported in the paper (e.g., preliminary or failed experiments that didn't make it into the paper). 
    \end{itemize}
    
\item {\bf Code of ethics}
    \item[] Question: Does the research conducted in the paper conform, in every respect, with the NeurIPS Code of Ethics \url{https://neurips.cc/public/EthicsGuidelines}?
    \item[] Answer: \answerYes{}
    \item[] Justification: All datasets used are existing public benchmarks employed under their original terms; the paper involves no human subjects, no scraped or sensitive personal data, and no model release with elevated misuse risk, and we have reviewed the NeurIPS Code of Ethics.
    \item[] Guidelines:
    \begin{itemize}
        \item The answer \answerNA{} means that the authors have not reviewed the NeurIPS Code of Ethics.
        \item If the authors answer \answerNo, they should explain the special circumstances that require a deviation from the Code of Ethics.
        \item The authors should make sure to preserve anonymity (e.g., if there is a special consideration due to laws or regulations in their jurisdiction).
    \end{itemize}

\item {\bf Broader impacts}
    \item[] Question: Does the paper discuss both potential positive societal impacts and negative societal impacts of the work performed?
    \item[] Answer: \answerYes{}
    \item[] Justification: Section~\ref{sec:conclusion} discusses positive impacts (faster, more reliable diagnosis of faults in safety-critical cyber-physical systems, reducing downtime and incident risk in water treatment, water distribution, and building management) and negative impacts (operator over-reliance on LLM rankings, automation bias in safety-critical loops, and miscalibrated confidence on out-of-distribution faults), together with mitigations (human-in-the-loop deployment, presenting the top $k$ candidates rather than a single answer, and surfacing per-candidate evidence rather than only the final rank).
    \item[] Guidelines:
    \begin{itemize}
        \item The answer \answerNA{} means that there is no societal impact of the work performed.
        \item If the authors answer \answerNA{} or \answerNo, they should explain why their work has no societal impact or why the paper does not address societal impact.
        \item Examples of negative societal impacts include potential malicious or unintended uses (e.g., disinformation, generating fake profiles, surveillance), fairness considerations (e.g., deployment of technologies that could make decisions that unfairly impact specific groups), privacy considerations, and security considerations.
        \item The conference expects that many papers will be foundational research and not tied to particular applications, let alone deployments. However, if there is a direct path to any negative applications, the authors should point it out. For example, it is legitimate to point out that an improvement in the quality of generative models could be used to generate Deepfakes for disinformation. On the other hand, it is not needed to point out that a generic algorithm for optimizing neural networks could enable people to train models that generate Deepfakes faster.
        \item The authors should consider possible harms that could arise when the technology is being used as intended and functioning correctly, harms that could arise when the technology is being used as intended but gives incorrect results, and harms following from (intentional or unintentional) misuse of the technology.
        \item If there are negative societal impacts, the authors could also discuss possible mitigation strategies (e.g., gated release of models, providing defenses in addition to attacks, mechanisms for monitoring misuse, mechanisms to monitor how a system learns from feedback over time, improving the efficiency and accessibility of ML).
    \end{itemize}
    
\item {\bf Safeguards}
    \item[] Question: Does the paper describe safeguards that have been put in place for responsible release of data or models that have a high risk for misuse (e.g., pre-trained language models, image generators, or scraped datasets)?
    \item[] Answer: \answerNA{}
    \item[] Justification: The paper releases no pre-trained model, no scraped data, and no generative artifact; the released code is a thin pipeline that orchestrates an existing third-party LLM API and existing public benchmarks, none of which carries a high misuse risk warranting additional safeguards.
    \item[] Guidelines:
    \begin{itemize}
        \item The answer \answerNA{} means that the paper poses no such risks.
        \item Released models that have a high risk for misuse or dual-use should be released with necessary safeguards to allow for controlled use of the model, for example by requiring that users adhere to usage guidelines or restrictions to access the model or implementing safety filters. 
        \item Datasets that have been scraped from the Internet could pose safety risks. The authors should describe how they avoided releasing unsafe images.
        \item We recognize that providing effective safeguards is challenging, and many papers do not require this, but we encourage authors to take this into account and make a best faith effort.
    \end{itemize}

\item {\bf Licenses for existing assets}
    \item[] Question: Are the creators or original owners of assets (e.g., code, data, models), used in the paper, properly credited and are the license and terms of use explicitly mentioned and properly respected?
    \item[] Answer: \answerYes{}
    \item[] Justification: We cite the original sources for every dataset (WADI~\cite{ahmed2017wadi}, SWaT~\cite{goh2016dataset}, HVAC/OEDI~\cite{OEDI_Dataset_5763}, RCAEval~\cite{pham2025rcaeval}) and every baseline implementation (BARO, RCD, $\epsilon$-Diagnosis, CIRCA, PageRank, RandomWalk) in Section~\ref{sec:experiments}, and Appendix~\ref{app:licenses} enumerates the license or terms of use of each dataset, code asset, and LLM backbone; package versions are listed in the released requirements file.
    \item[] Guidelines:
    \begin{itemize}
        \item The answer \answerNA{} means that the paper does not use existing assets.
        \item The authors should cite the original paper that produced the code package or dataset.
        \item The authors should state which version of the asset is used and, if possible, include a URL.
        \item The name of the license (e.g., CC-BY 4.0) should be included for each asset.
        \item For scraped data from a particular source (e.g., website), the copyright and terms of service of that source should be provided.
        \item If assets are released, the license, copyright information, and terms of use in the package should be provided. For popular datasets, \url{paperswithcode.com/datasets} has curated licenses for some datasets. Their licensing guide can help determine the license of a dataset.
        \item For existing datasets that are re-packaged, both the original license and the license of the derived asset (if it has changed) should be provided.
        \item If this information is not available online, the authors are encouraged to reach out to the asset's creators.
    \end{itemize}

\item {\bf New assets}
    \item[] Question: Are new assets introduced in the paper well documented and is the documentation provided alongside the assets?
    \item[] Answer: \answerYes{}
    \item[] Justification: The released code is under the MIT License and documented in its README, with one dataset registry and a command for each table (Appendix~\ref{app:reproducibility}); the per-dataset domain-knowledge documents are released with it, and the HVAC scenario construction is documented in Appendix~\ref{app:datasets}. No new datasets or models are released.
    \item[] Guidelines:
    \begin{itemize}
        \item The answer \answerNA{} means that the paper does not release new assets.
        \item Researchers should communicate the details of the dataset\slash code\slash model as part of their submissions via structured templates. This includes details about training, license, limitations, etc. 
        \item The paper should discuss whether and how consent was obtained from people whose asset is used.
        \item At submission time, remember to anonymize your assets (if applicable). You can either create an anonymized URL or include an anonymized zip file.
    \end{itemize}

\item {\bf Crowdsourcing and research with human subjects}
    \item[] Question: For crowdsourcing experiments and research with human subjects, does the paper include the full text of instructions given to participants and screenshots, if applicable, as well as details about compensation (if any)? 
    \item[] Answer: \answerNA{}
    \item[] Justification: The paper does not involve crowdsourcing or research with human subjects; all evaluation is conducted on existing public benchmarks containing only system telemetry.
    \item[] Guidelines:
    \begin{itemize}
        \item The answer \answerNA{} means that the paper does not involve crowdsourcing nor research with human subjects.
        \item Including this information in the supplemental material is fine, but if the main contribution of the paper involves human subjects, then as much detail as possible should be included in the main paper. 
        \item According to the NeurIPS Code of Ethics, workers involved in data collection, curation, or other labor should be paid at least the minimum wage in the country of the data collector. 
    \end{itemize}

\item {\bf Institutional review board (IRB) approvals or equivalent for research with human subjects}
    \item[] Question: Does the paper describe potential risks incurred by study participants, whether such risks were disclosed to the subjects, and whether Institutional Review Board (IRB) approvals (or an equivalent approval/review based on the requirements of your country or institution) were obtained?
    \item[] Answer: \answerNA{}
    \item[] Justification: The paper does not involve human subjects research, so IRB approval was not applicable.
    \item[] Guidelines:
    \begin{itemize}
        \item The answer \answerNA{} means that the paper does not involve crowdsourcing nor research with human subjects.
        \item Depending on the country in which research is conducted, IRB approval (or equivalent) may be required for any human subjects research. If you obtained IRB approval, you should clearly state this in the paper. 
        \item We recognize that the procedures for this may vary significantly between institutions and locations, and we expect authors to adhere to the NeurIPS Code of Ethics and the guidelines for their institution. 
        \item For initial submissions, do not include any information that would break anonymity (if applicable), such as the institution conducting the review.
    \end{itemize}

\item {\bf Declaration of LLM usage}
    \item[] Question: Does the paper describe the usage of LLMs if it is an important, original, or non-standard component of the core methods in this research? Note that if the LLM is used only for writing, editing, or formatting purposes and does \emph{not} impact the core methodology, scientific rigor, or originality of the research, declaration is not required.
    \item[] Answer: \answerYes{}
    \item[] Justification: An LLM is a core component of the proposed pipeline as the listwise reranker over the candidate set $\mathcal{C}$ in Section~\ref{sec:remedy}, Section~\ref{sec:experiments} documents the model identity, run protocol, and the no-DK and with-DK configurations, and Appendix~\ref{app:reproducibility} gives the prompt templates and decoding configuration, with second-backbone and temperature controls in Appendix~\ref{app:robustness}. A separate LLM from a model family disjoint from the evaluated backbones was used offline to draft the optional domain-knowledge documents and to rewrite them for the de-identification control.
    \item[] Guidelines:
    \begin{itemize}
        \item The answer \answerNA{} means that the core method development in this research does not involve LLMs as any important, original, or non-standard components.
        \item Please refer to our LLM policy in the NeurIPS handbook for what should or should not be described.
    \end{itemize}

\end{enumerate}
\end{document}